\documentclass[lettersize,journal]{IEEEtran}
\usepackage{amsmath,amsfonts}
\usepackage{algorithmic}
\usepackage{algorithm}
\usepackage{array}
\usepackage[caption=false,font=normalsize,labelfont=sf,textfont=sf]{subfig}
\usepackage{textcomp}
\usepackage{stfloats}
\usepackage{url}
\usepackage{verbatim}
\usepackage{graphicx}
\usepackage{cite}
\usepackage[backref]{hyperref}
\begin{document}

\title{Evaluating Similitude and Robustness of Deep Image Denoising Models via Adversarial Attack}

\author{Jie Ning, Jiebao Sun,Yao Li, Zhichang Guo, Wangmeng Zuo
\thanks{Jie Ning is with the School of Mathematics, Harbin Institute of Technology, Harbin, Heilongjiang, e-mail: 22B912021@stu.hit.edu.cn}
\thanks{Yao Li is with the School of Mathematics, Harbin Institute of Technology, Harbin, Heilongjiang, e-mail: yaoli0508@hit.edu.cn}
\thanks{Wangmeng Zuo is with the School of Computer Science and Technology, Harbin, Heilongjiang, e-mail: cswmzuo@gmail.com}}

\markboth{Journal of \LaTeX\ Class Files,~Vol.~14, No.~8, August~plug2023}%
{Shell \MakeLowercase{\textit{et al.}}: A Sample Article Using IEEEtran.cls for IEEE Journals}

\IEEEpubid{}

\maketitle

\begin{abstract}
Deep neural networks (DNNs) have shown superior performance comparing to traditional image denoising algorithms. However, DNNs are inevitably vulnerable while facing adversarial attacks. In this paper, we propose an adversarial attack method named denoising-PGD which can successfully attack all the current deep denoising models while keep the noise distribution almost unchanged. We surprisingly find that the current mainstream non-blind denoising models (DnCNN, FFDNet, ECNDNet, BRDNet), blind denoising models (DnCNN-B, Noise2Noise, RDDCNN-B, FAN), plug-and-play (DPIR, CurvPnP) and unfolding denoising models (DeamNet) almost share the same adversarial sample set on both grayscale and color images, respectively. Shared adversarial sample set indicates that all these models are similar in term of local behaviors at the neighborhood of all the test samples. Thus, we further propose an indicator to measure the local similarity of models, called robustness similitude. Non-blind denoising models are found to have high robustness similitude across each other, while hybrid-driven models are also found to have high robustness similitude with pure data-driven non-blind denoising models. According to our robustness assessment, data-driven non-blind denoising models are the most robust. We use adversarial training to complement the vulnerability to adversarial attacks. Moreover, the model-driven image denoising BM3D shows resistance on adversarial attacks.
\end{abstract}


\section{Introduction}
\IEEEPARstart{I}{mages} are prone to be contaminated from various noises during forming and transmission, which degrades the visual effect of the image and complicates the downstream tasks. Image noise degradation in optical imaging is usually described by:

\begin{equation}
	\label{noise}
	f=u+n, \quad u \in \Omega \text {, }
\end{equation}
where $\Omega $ is the image domain, $f$ is the observed image, $u$ is the clean image, and $n$ is zero mean noise usually assumed to follow a Gaussian distribution. The key task of image denoising is to recover $u$ from $f$. In particular, when noise information such as noise level and distribution are unknown, the task is called blind denoising. When the noise is known, it is called non-blind denoising. 

Image denoising methods can be broadly classified into three categories: model-driven methods, data-driven methods, and model-data hybrid-driven methods. Model-driven methods are physical meaning constrained and interpretable, such as TV\cite{TV}, PM\cite{PM}, BM3D\cite{BM3D}, therefore features disobey physical properties will not be generated. Model-driven methods work well in low-level noise sicarios. These model-driven methods work well in the case of low-level noise. However, they generally suffer from manual parameter and time-consuming optimization. Furthermore, incomplete manual priors are insufficient to express complex image structures, which limits their potential. In the past decade, data-driven methods based on deep neural networks have excellent performance, especially in the field of image denoising. According to the noise information demand, data-driven methods are divided into non-blind denoising methods (e.g. DnCNN\cite{DNCNN}, FFDNet\cite{FFDNET}, ECNDNet\cite{ECNDNet}, BRDNet\cite{BRDNet}) and blind denoising methods (e.g. DnCNN-B\cite{DNCNN}, Noise2Noise\cite{N2N}, RDDCNN-B\cite{RDDCNN}, FAN\cite{FAN}).  Data-driven methods rely on a large amount of input data to train the model in order to obtain optimal weight parameters to minimize the loss function. Powerful abilities to smooth out the noise and recover features are promised by data-driven methods, but the un-interpretability brought by their black-box structure is worrying. Model-data hybrid-driven methods are proposed to eliminate the shortcomings of artificially designed prior and the limitations of lack interpretability. There are two known types of hybrid-driven methods: plug-and-play and unfolding methods \cite{PPP,2017Learning,LPO}, such as DPIR\cite{DPIR}, CurvPnP\cite{li2022curvpnp}, and DeamNet\cite{DeamNet}. However, the performance of current hybrid-driven methods are still not as good as data-driven methods.

Model-driven image denoising methods are robust and flexible in terms of denoising range. While data-driven image denoising methods are less robust and more sensitive to noise variations. From the perspective of adversarial attacks, deep neural networks are susceptible to adversarial samples\cite{IP,EH}. By adding a carefully selected tiny perturbation to the input image, the system can make a wrong judgment, which can lead to serious consequences. This carefully designed method of generating small and imperceptible adversarial perturbations is known as adversarial attack\cite{7780651,RN21,RN20}. From a manifold learning perspective\cite{RN19}, high-dimensional data in nature are concentrated near nonlinear low-dimensional manifolds. Adversarial attacks can be considered as a malicious process that drags benign examples out of the manifolds where they are concentrated.

An interesting property of adversarial samples is their transferability, i.e., adversarial samples designed for one model can successfully fool another model. To the reason for the existence of adversarial samples transferability, there are many opinions but no conclusion has been reached yet. The prevailing claim is that the transferability is created due to the decision boundaries of the samples are correlated between different models\cite{F2017The,2017Universal,RN16}. This also suggests that the adversarial samples are not randomly generated, but are traceable subspaces in a high-dimensional space. Some literature\cite{2021Evolving,2016Transferability} attributes transferability to architectural similarity, where the transferability of adversarial samples between two models increases with architectural similarity. Some researches\cite{2021Disrupting,EH} believe that transferability is caused by the high linear correlation between features extracted in different deep neural networks. For the same set of samples, features extracted using one model can be correlated with features extracted using another model by a simple affine transformation. 

Existing adversarial attacks\cite{DN,EA} and defense techniques\cite{RDN,DA,DD} mostly target on classification tasks, while few involving image denoising tasks aim to pave the way for subsequent interference with other classification tasks\cite{cheng2021pasadena}. Genzel et al.\cite{9705105} argue that regression tasks, which mapping inputs to high-dimensional signal manifolds, are resilient to adversarial perturbations even under vanilla end-to-end network architectures. Furthermore, deep image denoising is an uninterpretable black-box model, so its stability, robustness, and the affective range of noise are unknown and rarely discussed. The same is true for the model-data hybrid-driven approach, which retains a portion of the network structure that brings uninterpretability, and whose stability, robustness, and the range of noise that can be removed remain to be verified.

Based on above reasons, we propose a PGD-based image denoising adversarial attack method, namely denoising-PGD, to disturb the regression task of deep image denoising. By adding adversarial perturbation to the input image, attacks are covertly embedded into the noise, i.e., the change in the noise distribution is imperceptible. The robustness of denoising models and the denoising mechanism of neural networks are explored from the performance variation. Moreover, since PGD-based adversarial attacks have low transferability among different model architectures, if adversarial samples generated by denoising-PGD exhibits strong transferability, high similitude between the models can be indicated. Our contribution consists of four aspects:

\noindent (1) We propose image denoising adversarial attack, a task for image denoising, with the aim of verifying the robustness of deep image denoising methods. The proposed denoising-PGD adversarial attack method has a significant adversarial effect and exhibits good generalization.

\noindent (2) Adversarial samples generated by denoising-PGD  shows excellent transferability in among various image denoising models, including non-blind denoising, blind denoising, plug-and-play, and unfolding models. The shared adversarial sample space is further experimentally explored which indicating that the current deep image denoising methods have high similitude.

\noindent (3) To improve the robustness of deep image denoising methods, adversarial training is conducted. It is found that adversarial training can significantly improve the visual effect in some pattern.   To explore this problem, a conjecture that adversarial noise constitutes a continuous space is proposed.

\noindent (4) The robustness of the classical model-driven method and the deep image denoising method on adversarial attacks is compared and analyzed. Classical model-driven methods BM3D shows some resistance to adversarial attack.

This paper is organized as follows. Section II reviews typical methods of image denoising and the adversarial attack method PGD. Section III describes the denoising-PGD method, its transferability among various denoising models, the exploration of the adversarial space, and the definition of robustness similitude. The experimental results are presented in Section IV. Section V analyzes and discusses the experimental results and presents some possible conjectures. Finally, Section VI concludes the paper.

\section{Related work}
\subsection{DnCNN and non-blind image denoising}
A milestone in data-driven models is DnCNN\cite{DNCNN} which implicitly recovers clean images through a residual learning strategy for predicting noise. The residual mapping will be easier to be optimized, when the original mapping is more like an identity mapping. By this characteristic, the residual learning with batch-normalization will improve the noise removal performance while stable the convergence. In terms of architecture design, DnCNN modifie the VGG convolutional filters and removes all pooling layers to increase the receptive field for utilizing more contextual information and making it suitable for image denoising. DnCNN greatly improves the image denoising task metrics, far surpassing the classical model-driven method BM3D.

FFDNet\cite{FFDNET} follows a similar framework but has an adjustable noise level map as input. The noise level map as an input allows the model parameters to not change with the noise level. Thus, FFDNet can be viewed as consisting of multiple denoisers with different noise levels, enabling control of the trade-off between denoising and detail retention. FFDNet denoising effect performs better in high noise cases and spatially varying noise than DnCNN. ECNDNet\cite{ECNDNet} continues the idea of DnCNN to solve the internal covariate shift problem and accelerate the network convergence. ECNDNet uses dilated convolution, which uses a dilated filter with a dilation factor to enlarge the receptive field for improving the performance. ECNDNet is more robust and more effective than commonly used denoising methods. BRDNet\cite{BRDNet} combines two networks and dilated convolution to increase the width of the network and the receptive field to obtain more features. Instead of BN operations that are ineffective in mini-batch problems, BRDNet uses BRN operations that approximate the distribution of the training data by a single sample rather than the entire mini-batch to prevent gradient problems and improve performance. BRDNet outperforms DnCNN and FFDNet in terms of image denoising metrics and is highly competitive.

\subsection{Blind denoising}
Zhang et al.\cite{DNCNN} introduced a DnCNN-based blind Gaussian denoising model, namely DnCNN-B. DnCNN-B gathers noise images from a wide range of noise levels to train a single model, thus it can perform denoising without knowing the noise level. DnCNN-B is no less impressive as a blind denoising method, with metrics levels close to DnCNN, and even outperforms DnCNN in the high noise case. Noise2Noise\cite{N2N} cleverly utilizes the statistical inference property of the l2-norm, that is, as long as the mean value of the input is constant, the valuation of the output remains constant, to achieve unsupervised blind denoising. Excellent denoising performance is obtained applying independent and identically distributed Gaussian noise images as both input and target. It verifies that using noise images yields the same performance as using clean target images as long as the noise mean is zero. RDDCNN-B\cite{RDDCNN} introduces deformable learnable kernels and stacked convolutional structure to extract more representative noise features based on the relationship of surrounding pixels. RDDCNN-B has a smaller number of parameters and outperforms DnCNN-B in both qualitative and quantitative analysis. As a blind model-data hybrid-driven method, FAN\cite{FAN} combines frequency domain analysis with attention mechanism. Its wavelet transform creates sparser features in frequency domain making full use of spectral information and spatial information. The goal of the neural network in FAN is to estimate the optimal solution of wavelet coefficients of clean images using nonlinear features. In the case of different noise distributions, FAN has a significant advantage over DnCNN and FFDNet in blind image denoising.

\subsection{Plug-and-play and unfolding}
These data-driven methods perform well, but often have undesired artificial effects and uninterpretability. As a result, model-data hybrid-driven methods emerged, such as plug-and-play and unfolding. The plug-and-play method is known for its flexibility and efficiency in handling various inverse problems. The main idea of this method is to expand the energy function into subproblems by using a variable splitting algorithm and then solving the subproblems associated with the prior using a pre-trained data-driven prior. In this approach, network training and model solving are separated. In contrast, unfolding integrates model-driven and data-driven approaches into an end-to-end network that performs network parameter optimization and model solving simultaneously. These model-data hybrid-driven methods have a complex mathematical foundation, which brings partial interpretability.

DPIR\cite{DPIR} trains a highly flexible and effective CNN denoiser and then inserts it as a module into a semi-quadratic splitting-based iterative algorithm, providing a prior that more closely fits the image structure. DPIR significantly improves the effectiveness of model-driven methods due to the powerful implicit a priori modeling of deep image denoising. In the absence of task-specific training, it is more flexible than data-driven methods such as DnCNN, while offering comparable performance. Li et al.\cite{li2022curvpnp} developed a plug-and-play blind denoising model CurvPnP, including a noise estimation sub-network to provide the noise level to the denoising sub-network. CurvPnP uses a depth curvature denoiser in the denoising module, which introduces Gaussian curvature mapping that contain rich fine structures and image details into the encoder-decoder architecture and supervised attention module as a prior. CurvPnP is more flexible to adapt to different image recovery tasks and outperforms state-of-the-art plug-and-play methods. DeamNet\cite{DeamNet} is a novel end-to-end and deep unfolding denoising network, which designed around adaptive consistency prior (ACP) and DEAM modules. By introducing nonlinear filtering operators, reliability matrices and high-dimensional feature transformation functions into the consistency prior, ACP can better portray the image information. Incorporates ACP into the maximum a posteriori framework and unfolding it to get the dual element-wise attention mechanism (DEAM) module for more feature. DeamNet improves the interpretability of the network to a certain extent and outperforms BRDNet in terms of denoising performance.

\textit{Summary and our finding:} In the last decade, deep learning has excelled in the field of image denoising, and the metrics have reached a very high level. Some scholars even believe that the problem in the direction of image denoising has been completely solved and does not need to spend too much effort on research. However, our study finds that all these excellent deep image denoising methods can be fooled, and even all of them can be easily fooled by the same image.

\subsection{PGD adversarial attack}
Goodfellow et al.\cite{EH} proposed an efficient one-step attack method based on the gradient of deep neural network, called the Fast Gradient Sign Method (FGSM).  FGSM uses relationship between the noise map and  image pixels as the loss function. The gradient of the loss function is computed by subtracting or adding a fixed value $\varepsilon $ from each pixel depending on its sign. The basic iterative method (BIM)\cite{RN39} replaces the one-step algorithm with multiple small steps, extending the FGSM to an iterative algorithm. Another iterative version of FGSM is the projection gradient descent (PGD) method\cite{RN17}, which uses projection gradient descent on a negative loss function to generate adversarial samples and can be described as a saddle point optimization problem. Unlike BIM, PGD perturbs the original image randomly before starting the iteration.

This saddle point optimization problem consists of an internal maximization problem and an external minimization problem. The internal maximization tries to find an adversarial sample of a given data point $x$ to maximize the loss, and the external minimization aims to find the model parameters that minimize the adversarial loss given by the internal maximization part. When using a loss function based on noise maps and image pixels, it is difficult to find local maxima that are significantly better than PGD. The authors of PGD verified in\cite{RN17} that if a network is trained to be robust to PGD adversarial samples, then it becomes robust to a wide range of other attacks as well. In the transferability study of adversarial samples, Lin et al.\cite{RN13} mentioned that typical gradient-based iterative attacks greedily perturb the image in the direction of the gradient sign at each iteration, usually falling into poor local maxima, so that have weak transferability. Because of this reason, PGD methods as an iterative version of FGSM are rarely mentioned in the study of adversarial attacks transferability. In summary, PGD is a concise and effective, representative adversarial attack method, so our study will be based on this method. In addition, as an adversarial attack method with low transferability in classification tasks, PGD that exhibits strong transferability in more complex and higher dimensional regression tasks indicates high similarity of the model.

Recently, Yan et al.\cite{2022adversarialdncnnb} proposed OBSATK, a two-step adversarial attack method for evaluating the robustness of deep denoising models, aiming to obtain a bounded perturbation that can decrease the quality of the denoised image. In the experiments, DnCNN-B were demonstrated to be vulnerable to adversarial attacks. Different from this work, we aim to explore characteristics of deep denoising models and the correlation between models from the adversarial attack perspective, rather than finding stronger attack methods. 

In this paper, we propose denoising-PGD which uses an additive operation to superimpose adversarial noise on a noise image, making it difficult for a trained deep image denoising model to perform the image denoising task properly. Since our adversarial noise distribution conforms to a Gaussian distribution, the composed noise should also be subordinate to Gaussian noise. Theoretically, noise images with added adversarial noise also fall into the denoising range of Gaussian denoising methods, and should not have difficulty in denoising. We try to find the reason of the excellent adversarial sample transferability among various image denoising methods by providing a conjecture of the similitude of deep image denoising tasks.

\section{Methodology}
\subsection{Motivation}
Deep image denoising is a regression task, which is quite different from classification tasks where adversarial attacks are usually studied. Existing denoising networks vary widely, but they have the same training paradigm, which include clean images as labels and clean images adding zero-mean Gaussian noise as input. Thus, we propose denoising-PGD which follows the standard PGD attack but using the negative l2-norm between the model output and the clean image as the loss function. The adversarial perturbation is applied to the noise input image to maintain the noise level and distribution. 

\begin{figure}[!t]
	\centering
	\includegraphics[width=1\linewidth]{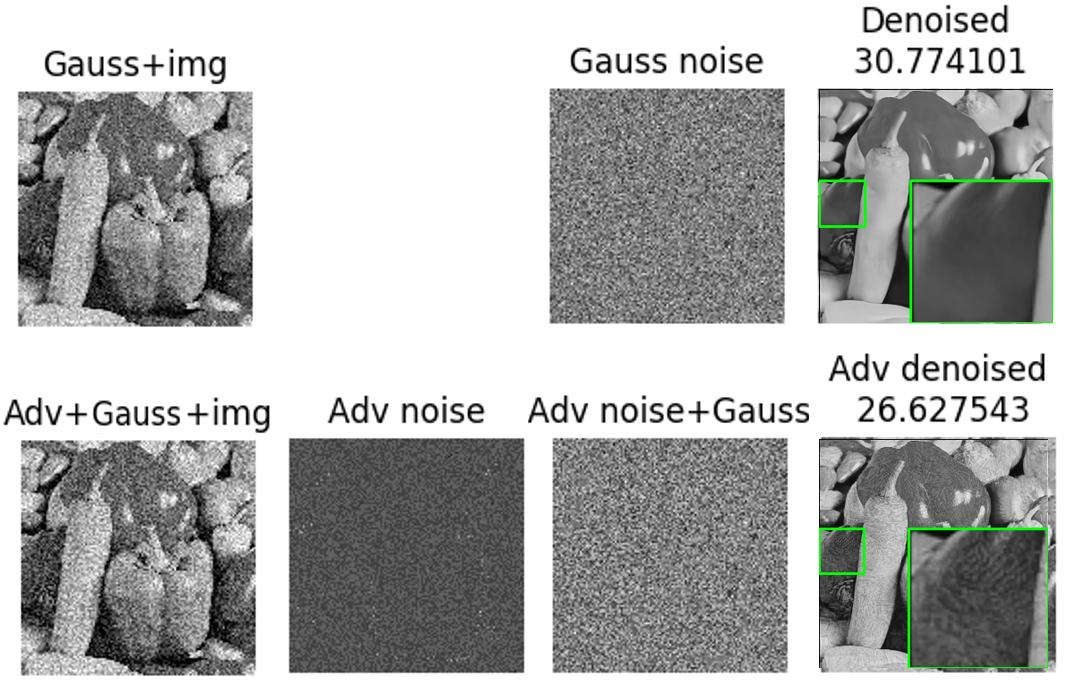}
	\caption{The performance of DnCNN decreases under adversarial attack.}
	\label{f:advshow}
\end{figure}

As seen in Fig. \ref{f:advshow}, the deep image denoising method has obvious adversarial effect on the adversarial sample. The adversarial noise is visually random and independent without obvious structural information. The magnitude of the adversarial noise is small and imperceivable. However, the PSNR of the attacked image after model denoising drops more than 4dB and obvious artifacts appear. Thus, the deep image denoising method is not robust in the tiny neighborhood around the Gaussian noise. In addition, the PGD-based adversarial attack is supposed to have poor transferability among different model structures. If the adversarial samples can easily transfer in different deep image denoising models, it will indicate that current deep image denoising models are essentially similar. To this end, we explore the robustness of different models and the adversarial attack transferability among different models.

\subsection{Image denoising adversarial attack}
\begin{figure}[!t]
	\centering
	\includegraphics[width=1\linewidth]{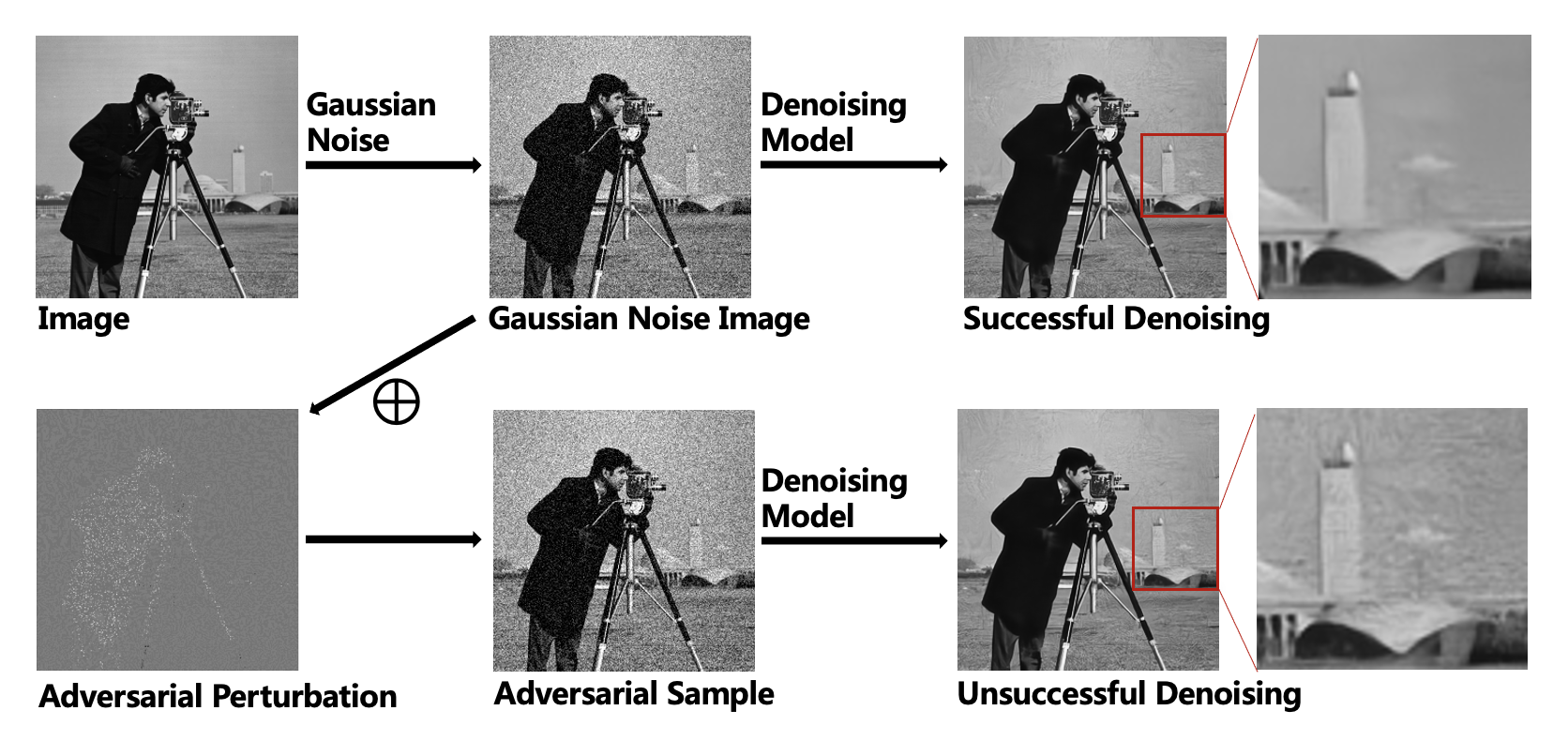}
	\caption{The flowchart of image denoising adversarial attack method. By adding small imperceptible adversarial perturbations to the noise image, the adversarial samples will fool the trained deep image denoising model and produce unsuccessful denoising results.}
	\label{f:flowchart}
\end{figure}

Given a noise image $x$, our goal is to generate an adversarial noise image ${x}' $ that can fool the deep image denoising model to result much lower denoising quality than functioning on a Gaussian noise image with the same noise level and similar distribution. We name this task as image denoising adversarial attack, whose flowchart is showed in Fig. \ref{f:flowchart}. 

For image denoising adversarial attacks, the following mathematical definition is given: let $D: R^{d} \rightarrow R^{d}$ be a denoiser, where $d$ is the image dimension. A noise image $x \in R^{d}$ is denoised as $E(D(x), \mathrm{y})$, where $y \in R^{d}$ is the clean image. The minimal adversarial sample $x^{\prime}\in R^{d}$ of $x$ w.r.t. a distance function $\gamma :R^{d} \rightarrow R_{+}$ is given as the solution  $x^{\prime} \in R^{d}$ of the optimization problem:

\begin{equation}
	\begin{array}{l}
		\min \gamma\left(x^{\prime}-x\right) \\
		\text { s.t. } E(D(x), y)-E\left(D\left(x^{\prime}\right), \mathrm{y}\right)>M
	\end{array}
\end{equation}

\noindent where $E$ is an arbitrary monotonically increasing function that evaluates the similarity between the original image and the predicted image, such as PSNR, SSIM or MAE. $M$ is a threshold value that measures a successful adversarial attack. In other words, $x^{\prime}$ is the closest point to $x$ with the distance function $\gamma$ whose denoising performance is lower than $x$.

\subsection{Image denoising adversarial attack transferability}
For the transferability of image denoising adversarial attacks, our study focuses on evaluating how the adversarial perturbations generated for the source denoising model $D_1$ affect the denoising results of the target denoising model $D_2$. Therefore, we choose the denoising effect of the target model on Gaussian noise as the baseline to compare the denoising effect of the target model on the adversarial samples with it, and consider the adversarial as successful if the effect decreases significantly. Therefore, we evaluate the transferability only for the adversarial samples to fool $D_1$, while the Gaussian noise images that were successfully denoised by deep image denoising model $D_1$ and $D_2$.

\begin{align}
	x^{\prime} \sim &O_{D_{1}, D_{2}}=\notag
	\\
	&	\left\{\begin{array}{c|c}
		& E\left(D_{1}(x), y\right)>E(x,y) \\
		x \sim O & E\left(D_{2}(x), y\right)>E(x,y) \\
		&E\left(D_{1}(x), y\right)-E\left(D_{1}\left(x^{\prime}\right), y\right)>M
	\end{array}\right\}
\end{align}

\noindent where the adversarial sample ${x}'=adv(x,D_1)$ is generated by an adversarial attack $adv(\cdot )$ on a Gaussian noise image $x$ in the original set $O$. For these adversarial samples, we say that the adversarial sample is transferable when $E\left(D_{2}(x), y\right)-E\left(D_{2}\left(x^{\prime}\right), y\right)>M$, and vice versa.

As shown in section IV.B, the adversarial samples generated by denoising-PGD have strong transferability in all the tested deep image denoising models. To further explore the latent reason, we take noise samples surrounding the Gaussian noise image. Specifically, we use denoising-PDG to generate adversarial perturbation and denote the perturbation as $v$. The Gaussian noise added to the clean image is denoted as $n$ and the clean image is denoted as $u$. The noise samples

\begin{equation}
	s=\left(\frac{n}{\|n\|} \sin \theta+\frac{v-<v, n>n}{\|v-<v, n>n\|} \cos \theta\right)\|v\|+n+u
\end{equation}
are collected on the circle centered at the Gaussian noise image and located in the plane whose basis are $n$ and $u$, where $\theta=\left\{\mathrm{k} / \mathrm{N}^{*} 2 \pi, \mathrm{k}=1, \ldots, \mathrm{N}\right\}$, and $<v, n>$ is the inner product of $v$ and $n$. Intuitively, we hope to observe the robustness of denoising models in a tiny sphere neighborhood of the Gaussian noise image. But as images are high-dimensional vectors, randomly collected samples may fail to collect the pivot behavior. Thus, we focus on the samples is a specific cross section of the sphere neighborhood, i.e. the circle in the $n,v$ plane. Note that, the direction of $n$ is set as the 0-angle of the circle. 

\begin{figure}[!t]
	\centering
	\includegraphics[width=0.9\linewidth]{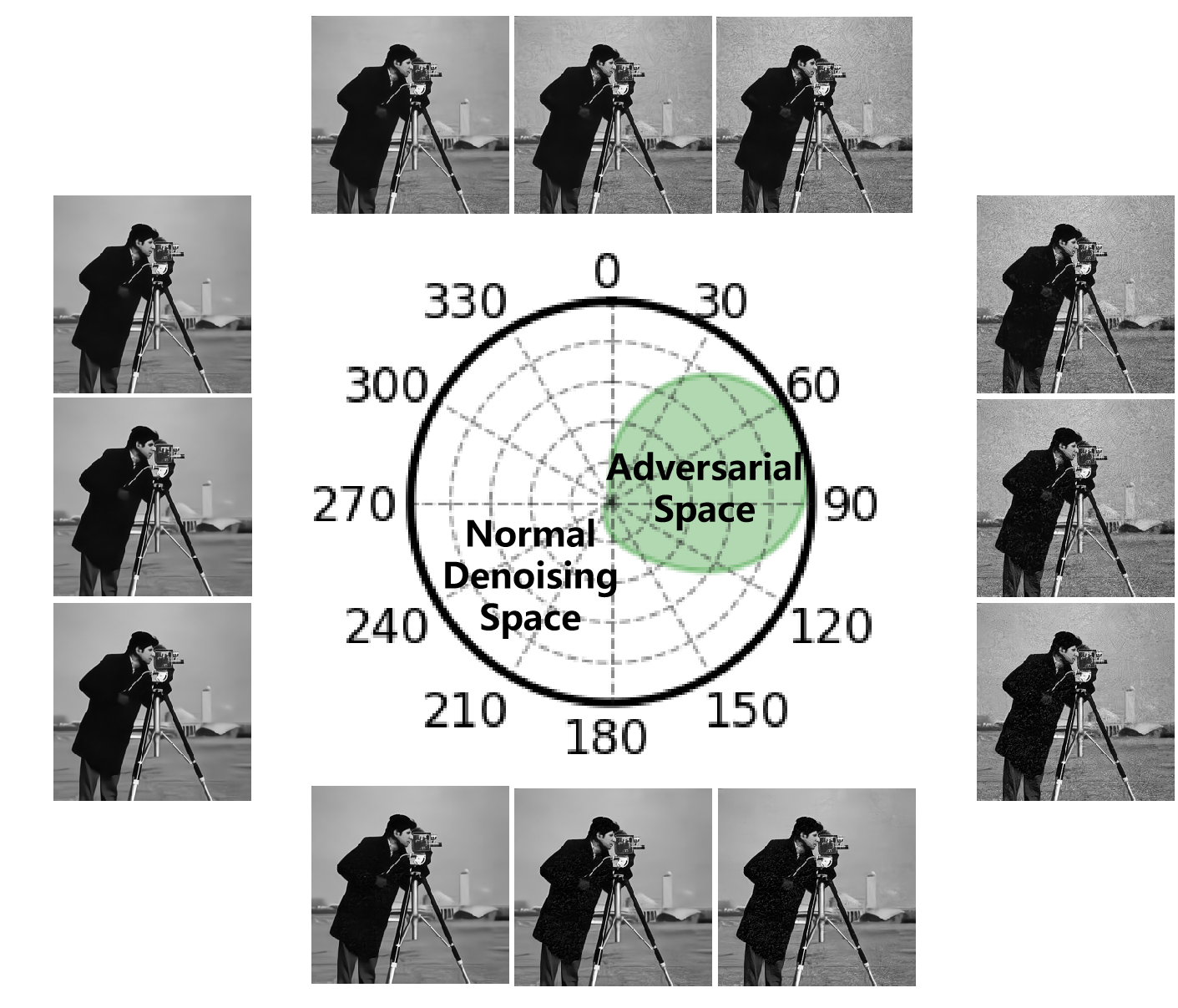}
	\caption{Radar diagram of the distribution of the adversarial space and normal denoising space. The closer green area to the edge the stronger is the adversarial performance. }
	\label{f:detail}
\end{figure}

\begin{figure}[!t]
	\centering
	\includegraphics[width=0.8\linewidth]{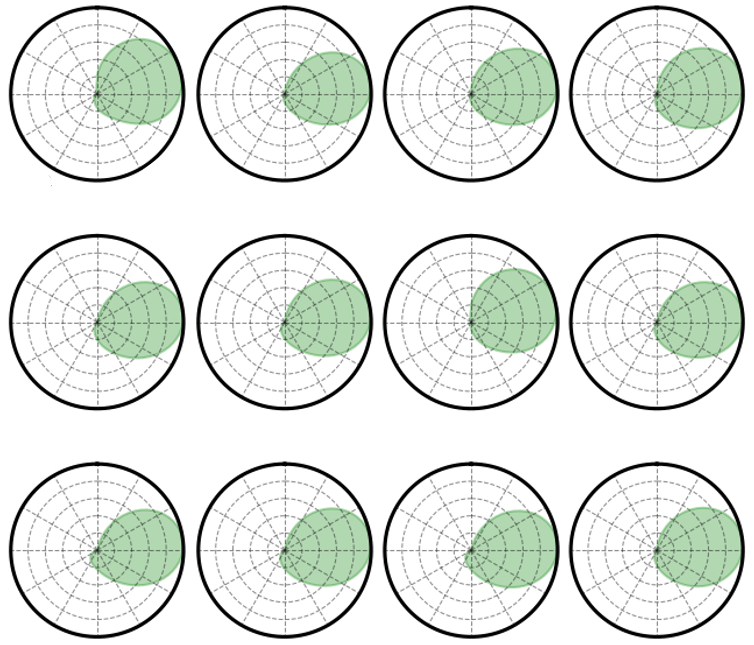}
	\caption{Adversarial regions of the DnCNN model under denoising-PGD attack on the dataset Set12.}
	\label{f:dncnn12}
\end{figure}

We generate noise samples on the classical non-blind deep denoising model DnCNN for the entire Set12 dataset with $N=100$. As shown in Fig. \ref{f:detail} and Fig. \ref{f:dncnn12}, the adversarial effect only happens on noise samples collected from 30 degrees to 150 degrees on the circle centered at the Gaussian noise “cameraman”. DnCNN presents performance drop and artifacts only on samples collected in the corresponding region. Surprisingly, we find that for all the images in Set12, the regions of adversarial samples are almost the same. In other word, for all the test images in Set12, the adversarial effect happens on noise samples collected from 30 degrees to 150 degrees. The interesting phenomenon indicates that the robustness of DnCNN is actually image content invariant, as the adversarial space remains the same for the entire test dataset.

\begin{figure}[!t]
	\centering
	\includegraphics[width=1\linewidth]{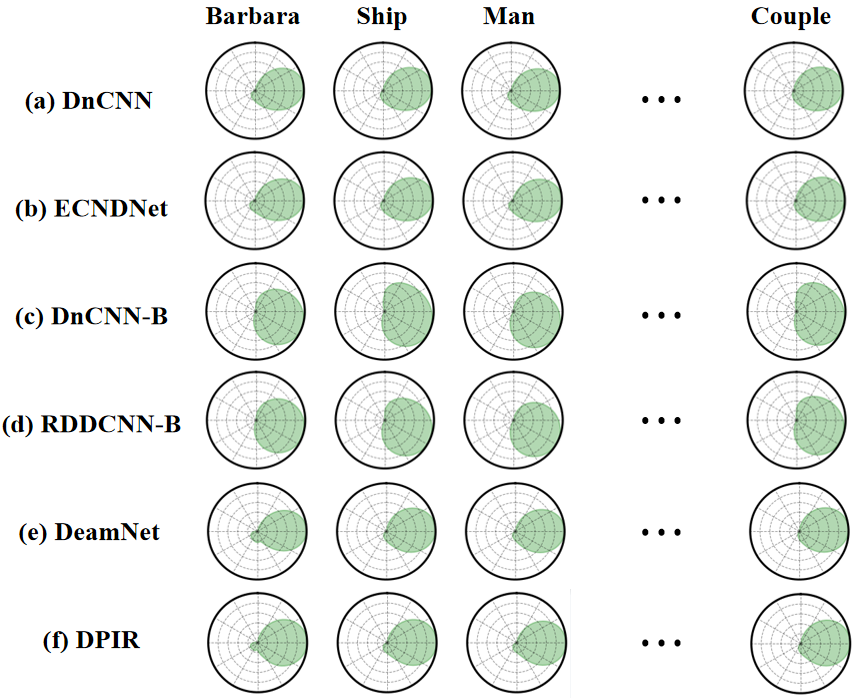}
	\caption{Adversarial regions of various models under adversarial samples generated for DnCNN.}
	\label{f:alladvrange}
\end{figure}

In addition, we tested these generated noise samples on the other deep image denoising models, such as the non-blind denoising model ECNDNet, blind denoising models DnCNN-B and RDDCNN-B,  the unfolding model DeamNet, and the plug-and-play model DPIR. The adversarial behaviors of these models are observed and recorded in Fig. \ref{f:alladvrange}. For all the tested models, the adversarial regions are concentrated in the angular range from 30 degrees to 150 degrees. It means that all the tested models regressed similar denoising functions which leads to similar adversarial regions. It also explains why the denoising-PGD has unexpected excellent transferability among different types of denoising models, as their adversarial spaces are almost overlap with each other. In a deeper observation, the non-blind denoising model (a) and (b) perform similarly, and the blind denoising model (c) and (d) perform similarly. This demonstrates that the adversarial ranges are similar on the same type of deep image denoising networks. The model (e) is the unfolding method and (f) is the plug-and-play method. Although both of them are model-data hybrid-driven methods, their adversarial regions are similar to pure data-driven methods. At a closer look, DPIR has larger adversarial regions.

\subsection{Denoising model robustness and robustness similitude}

We evaluate the denoising model robustness (DMR) as
\begin{align}
	DMR = \frac{1}{M} \sum_{i=1}^M  E(D(x_i^\prime))-E(D(x_i))
\end{align}
where $X=\{x_1,\hdots,x_M\}$ is the evaluation dataset and $x_i^\prime=adv(x_i, D)$ is the adversarial sample. DMR calculates the average metric change caused by applying adversarial attack to the target model on a specified dataset.

As demonstrated in the previous subsection, all the tested models regress a similar denoising function or at least similar in the local function regression around all the test data.
To evaluate the similarity across models, or precisely the local similarity around all the test data, we propose an indicator called denoising model robustness similitude (DMRS):
\begin{align}
	DMRS = \frac{1}{M} \sum_{i=1}^M \text{IoU}(\text{Radar}(D_1(s_i)), \text{Radar}(D_2(s_i)))
\end{align}
where $\text{IoU}$ is the Intersection over Union, $\text{Radar}$ is the radar diagram calculated on the $n,v$-plane cross section of the sample neighborhood, i.e. 
\begin{align}
	\text{Radar}(D(s_i)) = \text{enclosed area of } \left\{ \frac{E(D(s_{ij}))}{\max_j E(D(s_{ij}))}\right\}
\end{align}
where $s_{ij}$ is the noise sample of $x_i$ at $\theta_j = j/N*2\pi$. Intuitively, DMRS is the averaged IoU between two different models $D_1$ and $D_2$ over the evaluation dataset. High DMRS indicates two models are essentially similar observed from the robustness perspective.

\subsection{Image denoising adversarial space}
According to above observations, adversarial regions are similar across models and across input images. To verify that the adversarial effect is almost everywhere in the region instead of isolated points, we analyzed the linear combination of any two adversarial samples obtained in the noise sample circle. The linear combination is

\begin{equation}
	\mathrm s_l=\lambda\mathrm s_1+(1-\lambda)\mathrm s_2
\end{equation}
where $\lambda\in \left [ 0,1 \right ] $ is the hyperparameter. $s_1,s_2$ are the adversarial instances and $s_l$ is the new example formed by the combination.

\begin{figure}[!t]
	\centering
	\includegraphics[width=1\linewidth]{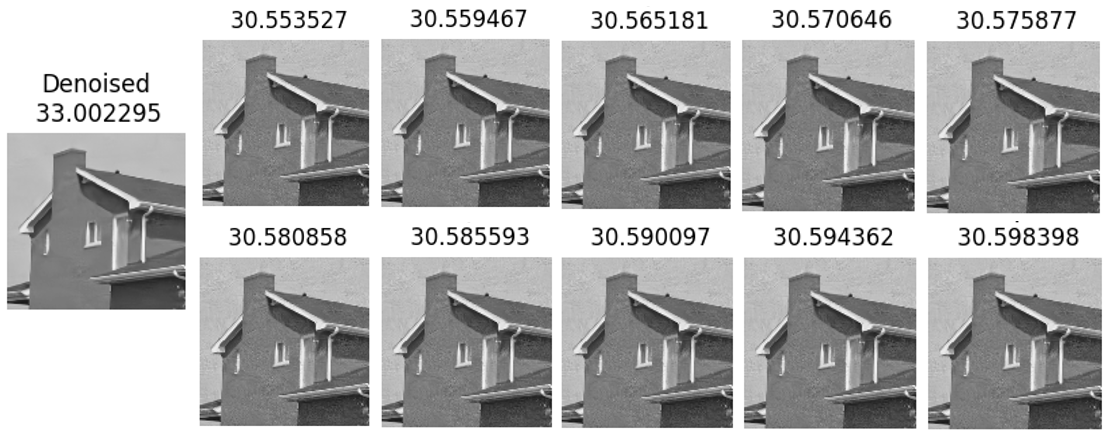}
	\caption{A linear combination of two arbitrary adversarial samples, where the coefficients are traversed to 1 in steps of 0.1.}
	\label{f:combineadv}
\end{figure}

In Fig. \ref{f:combineadv}, it can be seen that the linear combination of two adversarial samples can generate new adversarial samples. In other words, the image denoising adversarial region constitutes a continuous space in which an infinite number of adversarial samples lurk.

\section{Experiments}
\textit{Dataset}: We follow the setup of [4] using CImageNet400 dataset for training, which has 400 grayscale images of size 180 × 180 .  We embedding adversarial attacks in the testing phase, and use the Set12 and Set3c dataset for effect evaluation. These images are widely used for the evaluation of Gaussian denoising methods and they are not included in the training dataset.

\textit{Metrics}: There are two widely used metrics for full-reference image quality assessment: peak signal-to-noise ratio (PSNR) and structural similarity (SSIM) \cite{ZHOU2019102655}: 

\begin{align}
	\operatorname{PSNR}(\mathbf{X}, \mathbf{Y}) & =20 \cdot \log _{10}\left(\frac{\operatorname{MAX}_{\mathrm{I}}}{\operatorname{MSE}(\mathbf{X}, \mathbf{Y})}\right) \\
	\operatorname{SSIM}(\mathbf{X}, \mathbf{Y}) & =\frac{\left(2 \mu_{x} \mu_{y}+c_{1}\right)\left(2 \sigma_{x y}+c_{2}\right)}{\left(\mu_{x}^{2}+\mu_{y}^{2}+c_{1}\right)\left(\sigma_{x}^{2}+\sigma_{y}^{2}+c_{2}\right)}
\end{align}
where $X$ and $Y$ are the two images being used for comparison. $\mu _x$, $\mu _y$, $\sigma _{x}^{2} $, $\sigma _{y}^{2} $ are corresponding mean variance values, and $\sigma _{xy}$ is the covariance. $MAX_I$ is the maximum intensity, which is usually 255 in 8-bit representation.

To evaluate the effectiveness of adversarial denoising attacks, in addition to these two metrics, we chose the mean absolute error (MAE) to measure the performance of image denoising adversarial attack methods.

\begin{equation}
	\text { MAE }=\frac{\|X-Y\|_{1}}{N_{x} N_{y}}
\end{equation}
where $N_X$, $N_Y$ are the image size. MAE provides a good point-by-point measure of the error.

\subsection{The denoising-PGD for image denoising adversarial attack}
We apply denoising-PGD on DnCNN model to generate adversarial samples for Set12. The noise level set to 25, maximum perturbation $\varepsilon$ set to 0.012, and the number of iteration steps is set to 5.

\begin{figure}[!t]
	\centering
	\includegraphics[width=1\linewidth]{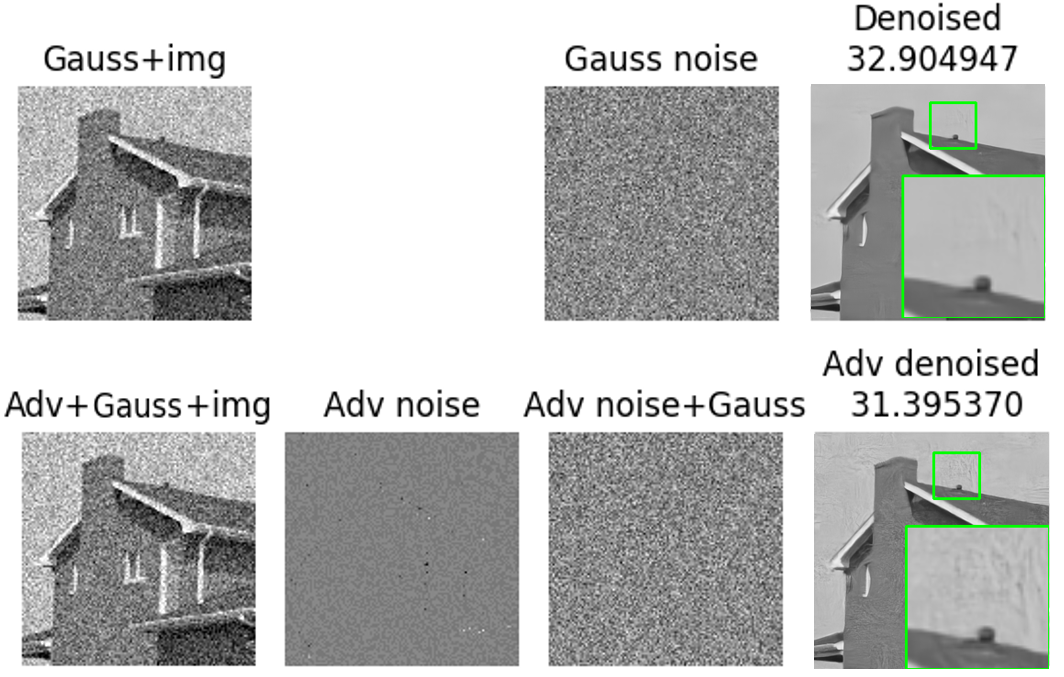}
	\caption{The denoising-PGD adversarial attack effect (house).}
	\label{f:dncnnadv}
\end{figure}

The adversarial noise we generate is a perturbation of very small noise level superimposed on a Gaussian noise image. However, the tiny perturbation can nevertheless produce a great disturbance in the denoising process. As shown in Fig. \ref{f:dncnnadv}, the adversarial noise image is visually the same as the Gaussian noise image, but the denoising effect shows a great difference, with a PSNR difference of 1.510 dB. The magnified area in the image shows that the artifacts that are slight in the Gaussian denoised image are very serious in the adversarial denoised image, and the texture details of the image are also changed.

Table \ref{ta:dncnnaadv} lists the adversarial effects caused by denoising-PGD on DnCNN for Set12. The adversarial sample of each image behaves significantly worse than the Gaussian noise image during the denoising process in every evaluation metrics. In average, PSNR drops up to 0.8 dB, SSIM drops by 0.04, and MAE increases by 0.8.

\begin{table}
\caption{Comparison of the denoising effect on Gaussian noise images and adversarial noise images for DnCNN model.}
\fontsize{7}{9}\selectfont
\centering
\begin{tabular}{ p{1cm}p{.5cm}p{1cm}p{.5cm}p{1cm}p{.5cm}p{1.2cm} } 
	\hline
	\hline
	& Gauss & Adv & Gauss & Adv & Gauss & Adv \\
	Image & PSNR & PSNR(RES) & MAE & MAE(RES) & SSIM & SSIM(RES) \\
	\hline
	Cameraman & 30.0 & 29.1(-0.9) & 5.1 & 6.3(+1.2) & 0.87 & 0.81(-0.06) \\
	House & 33.0 &31.4(-1.6) & 4.0 & 5.1(+1.1) & 0.86 & 0.79(-0.07) \\
	Pepper & 30.7 &29.6(-1.1) & 5.3 & 6.2(+0.9) & 0.88 & 0.82(-0.06) \\
	Fishstar & 29.4 &28.7(-0.7) & 6.3 & 7.0(+0.7) & 0.87 & 0.85(-0.02) \\
	Monarch & 30.3 &29.6(-0.7) & 5.3 & 6.0(+0.7) & 0.92 & 0.87(-0.05) \\
	Airplane & 29.1 &28.6(-0.5) & 5.8 & 6.4(+0.6) & 0.87 & 0.84(-0.03) \\
	Parrot & 29.5 &28.7(-0.8) & 5.8 & 6.7(+0.9) & 0.86 & 0.81(-0.05) \\
	Lena & 30.9 &29.9(-1.0) & 5.0 & 6.0(+1.0)& 0.88 & 0.82(-0.06) \\
	Barbara & 28.5 &28.1(-0.4) & 6.9 & 7.3(+0.4)& 0.83 & 0.82(-0.01) \\
	Ship & 29.0 &28.3(-0.7) & 6.4 & 7.2(+0.8)& 0.83 & 0.79(-0.04) \\
	Man & 28.8 &28.2(-0.6) & 6.7 & 7.4(+0.7)& 0.81 & 0.78(-0.03) \\
	Couple& 28.7 &27.9(-0.8) & 6.8 & 7.7(+0.9)& 0.83 & 0.79(-0.04) \\
	\hline
	Avg. & 29.8 &29.0(-0.8) & 5.8 & 6.6(+0.8)& 0.86 & 0.82(-0.04) \\
	\hline
	\hline
\end{tabular}
\label{ta:dncnnaadv}
\end{table}

Considering that the adversarial noise is superimposed on the original Gaussian noise image in this paper, this operation may bring a noise level shift to the noise image. To ensure the experimental rigor, we generated a Gaussian noise map with the same noise level with reference to the adversarial noise image to test the denoising ability of the deep image denoising method. It is found that the deep image denoising method can denoise such Gaussian noise images with slightly fluctuating noise levels normally. Therefore, it is verified that our adversarial sample is not adversarial due to perturbed noise levels.

Further, we use denoising-PGD to generate adversarial samples for a blind denoising model DnCNN-B with the same hyperparameter setting. In Table \ref{ta:dncnnbadv}, it shows that the PSNR drop is up to 4.0 dB, the SSIM drop is 0.22, and the MAE difference is 4.1 on DnCNN-B. The adversarial effect is much stronger on DnCNN-B and the visually difference is much obvious with many noise traces remaining on the image. DnCNN-B is less robust comparing to DnCNN.

\begin{table}
\caption{Comparison of the denoising effect on Gaussian noise images and adversarial noise images for DnCNN-B model.}
\fontsize{7}{9}\selectfont
\centering
\begin{tabular}{ p{1cm}p{.5cm}p{1cm}p{.5cm}p{1cm}p{.5cm}p{1.2cm} } 
	\hline
	\hline
	& Gauss & Adv & Gauss & Adv & Gauss & Adv \\
	Image & PSNR & PSNR(RES) & MAE & MAE(RES) & SSIM & SSIM(RES) \\
	\hline
	Cameraman & 30.0 & 25.1(-4.9) & 5.1 & 10.4(+5.3) & 0.87 & 0.56(-0.31)\\
	House & 32.9 &27.1(-5.8) & 4.0 & 8.4(+4.4) & 0.86 & 0.58(-0.28) \\
	Pepper & 30.7 &26.1(-4.6) & 5.3 & 9.5(+4.2) & 0.88 & 0.63(-0.25) \\
	Fishstar & 29.3 &25.4(-3.9) & 6.3 & 10.3(+4.0) & 0.87 & 0.70(-0.17) \\
	Monarch & 30.4 &27.2(-3.2) & 5.2 & 8.2(+3.0) & 0.92 & 0.74(-0.18) \\
	Airplane & 29.1 &25.4(-3.7) & 5.8 & 10.3(+4.5) & 0.87 & 0.61(-0.26) \\
	Parrot & 29.4 &24.7(-4.7) & 5.8 & 11.1(+5.3) & 0.86 & 0.57(-0.29) \\
	Lena & 30.8 &26.3(-4.5) & 5.1 & 9.2(+4.1)& 0.88 & 0.64(-0.24) \\
	Barbara & 28.3 &25.8(-2.5) & 7.0 & 9.7(+2.7)& 0.83 & 0.71(-0.12) \\
	Ship & 29.0 &25.6(-3.4) & 6.4 & 10.4(+4.0)& 0.83 & 0.64(-0.19) \\
	Man & 28.8 &28.2(-0.6) & 6.7 & 10.2(+3.5)& 0.81 & 0.65(-0.16) \\
	Couple& 28.7 &25.1(-3.6) & 6.8 & 10.8(+4.0)& 0.83 & 0.65(-0.18) \\
	\hline
	Avg.& 29.8 &25.8(-4.0) & 5.8 & 9.9(+4.1)& 0.86 & 0.64(-0.22) \\
	\hline
	\hline
\end{tabular}
\label{ta:dncnnbadv}
\end{table}

\subsection{Denoising model robustness evaluation}
We use denoising-PGD to generate adversarial samples against the target model on Set12. PSNR is used as evaluation metric, as shown in Table \ref{ta:robustness}. We can see that data-driven non-blind denoising models are the most robust, followed by hybrid-driven models, while blind denoising models are the least robust.

\begin{table}\label{Ta4}
	\caption{Denoising model robustness on Set12.}
	\fontsize{7}{10}\selectfont
	\centering
	\begin{tabular}{ p{1.3cm}p{.58cm}p{.58cm}p{.58cm}p{.58cm}p{.58cm}p{.58cm}p{.58cm} } 
		\hline
		\hline
		& &Gauss & Adv & Gauss & Adv & Gauss & Adv \\
		Image & DMR & PSNR & PSNR & MAE & MAE & SSIM & SSIM \\
		\hline
		DnCNN          & -0.94 & 30.40 & 29.47 & 5.39 & 6.28 & 0.862 & 0.819\\
		ECNDNet          & -0.87 & 30.38 & 29.51 & 5.40 & 6.25 & 0.861 & 0.814 \\
		DnCNN-B    & -4.53 & 30.31 & 25.78 & 5.44 & 9.91 & 0.858 & 0.610 \\
		RDDCNN-B                  & -3.80 & 30.39 & 26.50 & 5.46 & 8.96 & 0.859 & 0.665 \\
		DPIR           & -2.14 & 30.94 & 28.80 & 4.99 & 7.05 & 0.873 & 0.793 \\
		DeamNet                 & -1.27 & 30.80 & 29.53 & 5.12 & 6.32 & 0.871 & 0.799 \\
		\hline
		\hline
	\end{tabular}
\label{ta:robustness}
\end{table}

\subsection{The transferability of denoising-PGD}
To measure the transferability of generated adversarial samples, several denoising methods are selected for comparative and exploration, including non-blind denoising methods FFDNet and ECNDNet, blind denoising methods DnCNN-B and RDDCNN-B, the plug-and-play method DPIR and the unfolding method DeamNet. We generate adversarial samples on DnCNN and then test these adversarial samples on the above models.

\begin{figure}[!t]
	\centering
	\includegraphics[width=1\linewidth]{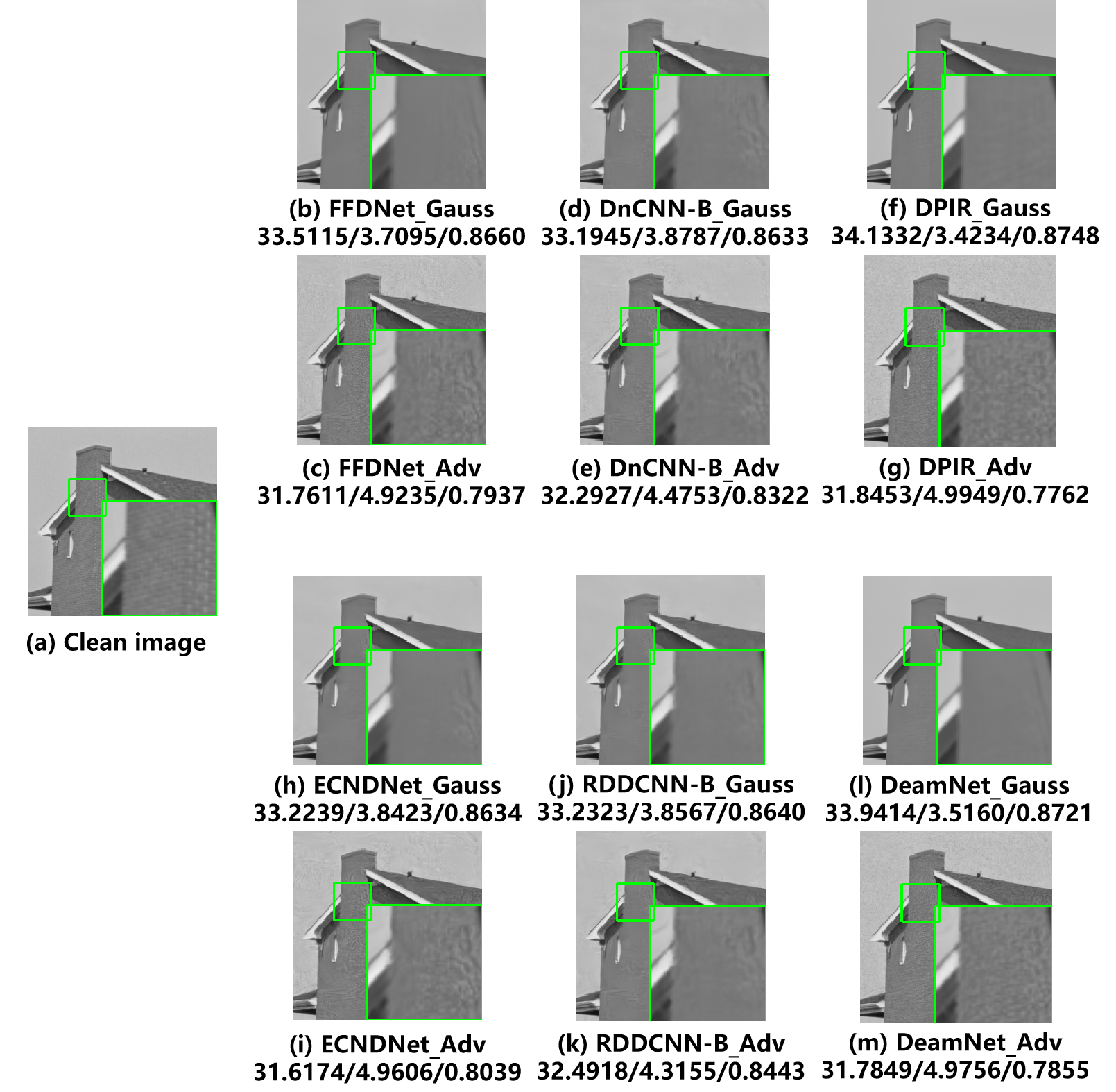}
	\caption{Transferability of adversarial samples generated by DnCNN to various models (house). The test values are PSNR/MAE/SSIM from left to right.}
	\label{f:dncnntran}
\end{figure}

We surprisingly find that all the tested models shown adversarial effect on adversarial samples generated for DnCNN. In the zoomed areas of Fig. \ref{f:dncnntran}, the Gaussian noise image after denoising exhibits some unrealistic smoothness while the adversarial noise image exhibits difficulty in removing the noise but produces blurred noise patches. From the measured metrics, there is a significant decrease in PSNR, down by 1.7504 dB on the FFDNet model, a lesser decrease of 0.9018 dB on the DnCNN-B model, and a largest decrease of 2.2879 dB on the DPIR model. In addition, the PSNR decreases by 1.6065 dB on ECNDNet, 0.7405 dB on RDDCNN-B, and 2.1565 dB on DeamNet.

We show the denoising performance of the adversarial samples generated on the DnCNN model using denoising-PGD on the Set12 dataset at noise level 25 on the non-blind denoising method FFDNet, the blind denoising method DnCNN-B, and the plug-and-play method DPIR in Table \ref{ta:dncnntran}. For the averaged PSNR, FFDNet decreases by 0.7 dB, DnCNN-B decreases by 0.7 dB, and DPIR decreases by 0.9 dB. The SSIM and MAE metrics show similar trends to each other. It can be seen that the blind denoising model is relative resistant to the adversarial perturbation created for the non-blind denoising model. However, the plug-and-play model, as a model-data hybrid-driven method, is unexpectedly sensitive to the adversarial perturbation for the non-blind denoising model. Note that, denoising-PGD, which is supposed to have poor transferability, exhibits excellent transferability among multiple deep image denoising methods which are designed in various motivations, mathematical backgrounds, and across years. It indicates that current deep denoising models are essentially similar, at least similar in the local function regression around all the test data.

\begin{table*}
\caption{Transferability of adversarial samples generated by DnCNN to FFDNet, DnCNN-B, and DPIR on Set12.}
\fontsize{7}{9}\selectfont
\centering
\begin{tabular}{ llllllllllllll|l } 
	\hline
	\hline
	Image &        &Cameraman&House&Pepper&Fishstar&Monarch&Airplane&Parrot&Lena&Barbara&Ship&Man&Couple&Avg.(RES)\\
	\hline
	&&&&&&&FFDNet&&&&&&&\\
	\hline
	           &PSNR&29.8&33.5&30.9&29.2&30.1&29.0&29.1&31.0&28.7&29.0&28.8&28.8&29.8 \\
	Gauss &SSIM&0.87&0.87&0.88&0.86&0.92&0.87&0.85&0.89&0.84&0.83&0.81&0.84&0.86 \\
	           &MAE&5.4&3.7&5.1&6.4&5.3&5.9&6.1&4.9&6.6&6.5&6.7&6.7&5.8\\
	           &PSNR&29.2&31.8&29.9&28.7&29.4&28.4&28.8&30.1&28.3&28.2&28.2&28.1&29.1(-0.7)\\
	Adv     &SSIM&0.80&0.79&0.82&0.85&0.87&0.83&0.81&0.84&0.83&0.78&0.78&0.80&0.82(-0.4)\\
	           &MAE&6.2&4.9&6.0&6.9&6.2&6.5&6.6&5.7&7.2&7.4&7.4&7.5&6.6(+0.8)\\
	\hline
	&&&&&&&DnCNN-B&&&&&&&\\
	\hline
	           &PSNR&29.1&33.2&30.4&29.1&30.2&28.9&28.5&30.9&28.4&28.8&28.8&28.6&29.6 \\
	Gauss &SSIM&0.83&0.86&0.87&0.86&0.91&0.86&0.83&0.88&0.83&0.82&0.81&0.83&0.85 \\
	           &MAE&5.9&3.9&5.5&6.5&5.3&5.9&6.6&5.1&6.9&6.6&6.8&6.9&6.0\\
	           &PSNR&28.0&32.3&29.5&28.2&29.9&28.3&27.3&30.4&28.1&28.3&28.4&28.1&28.9(-0.7)\\
	Adv     &SSIM&0.76&0.83&0.82&0.83&0.89&0.81&0.75&0.86&0.81&0.80&0.79&0.81&0.81(-0.4)\\
	           &MAE&6.8&4.5&6.2&7.3&5.7&6.7&7.8&5.4&7.3&7.1&7.2&7.4&6.6(+0.6)\\
	\hline
		&&&&&&&DPIR&&&&&&&\\
	\hline
		   &PSNR&30.1&34.1&31.2&29.8&30.7&29.4&29.3&32.4&29.9&29.2&29.0&29.1&30.3 \\
	Gauss &SSIM&0.87&0.87&0.89&0.88&0.93&0.87&0.86&0.90&0.88&0.84&0.82&0.85&0.87 \\
	           &MAE&5.3&3.4&5.0&6.0&4.9&5.5&6.0&4.6&5.8&6.2&6.5&6.4&5.5\\
	           &PSNR&29.3&31.8&29.9&29.3&29.7&28.7&28.8&30.2&29.2&28.5&28.4&28.4&29.4(-0.9)\\
	Adv     &SSIM&0.78&0.78&0.81&0.86&0.86&0.83&0.80&0.83&0.85&0.79&0.78&0.81&0.81(-0.6)\\
	           &MAE&6.3&5.0&6.1&6.4&6.0&6.2&6.7&5.7&6.5&7.1&7.2&7.2&6.4(+0.9)\\
	\hline
	\hline
\end{tabular}
\label{ta:dncnntran}
\end{table*}

In the experiments to test the transferability from DnCNN-B, we selected the non-blind denoising methods FFDNet and ECNDNet, the blind denoising methods DnCNN-B and RDDCNN-B, the plug-and-play method DPIR and the deep unfolding method DeamNet to test the adversarial samples generated by DnCNN-B under denoising-PGD. The results are compiled in Table \ref{ta:dncnnbtran}.

Comparing to the non-blind denoising method and plug-and-play method with less than 1 dB decrease in PSNR, the blind denoising method DnCNN-B decreases by 4.01 dB and RDDCNN-B decreases by 2.61 dB. The SSIM and MAE metrics also show the same trend, i.e., the adversarial attack samples generated with the blind denoising model DnCNN-B are highly transferable on the blind denoising model and is also transferable on non-blind denoising methods as well as plug-and-play methods, but the attack effect is largely weakened. A possible reason is that blind denoising methods have larger adversarial space comparing to non-blind denoising and plug-and-play methods. Thus, adversarial samples generated for DnCNN-B can fall out of the adversarial regions of non-blind denoising and plug-and-play methods.

\begin{table}
\caption{Transferability of adversarial samples generated by DnCNN-B to various models on Set12}
\fontsize{7}{9}\selectfont
\centering
\begin{tabular}{ p{1.3cm}p{.7cm}p{.7cm}p{.7cm}p{.7cm}p{.7cm}p{.7cm} } 
	\hline
	\hline
	& Gauss & Adv & Gauss & Adv & Gauss & Adv \\
	Image & PSNR & PSNR & MAE & MAE & SSIM & SSIM \\
	\hline
	FFDNet      & 29.66 & 29.73 & 5.88 & 5.89 & 0.858 & 0.858\\
	ECNDNet    & 29.74 & 29.63 & 5.86 & 5.98 & 0.858 & 0.854 \\
	DnCNN-B    & 29.78 & 25.77 & 5.80 & 9.88 & 0.857 & 0.641 \\
	RDDCNN-B & 29.60 & 26.99 & 5.96 & 8.40 & 0.852 & 0.717 \\
	DPIR           & 30.25 & 30.15 & 5.46 & 5.58 & 0.871 & 0.866 \\
	DeamNet    & 30.11 & 29.96 & 5.60 & 5.76 & 0.868 & 0.861 \\
	\hline
	\hline
\end{tabular}
\label{ta:dncnnbtran}
\end{table}

\subsection{Denoisng model robustness similitude evaluation}
We use denoising-PGD to generate adversarial samples on the Set12 dataset for the DnCNN model and then calculate corresponding noise samples at different angles as the evaluation noise sample set for the similitude evaluation of all the models. As shown in Table \ref{ta:similitude}, deep image denoising models of the same type have up to 90\% similarity even though the comparison models were proposed across many years. And surprisingly, the model-data hybrid-driven models with partial interpretability also has around 90\% similarity with the non-blind denoising models.

\begin{table}
	\caption{Denoising model robustness similitude on Set12 (DnCNN adv. sample based).}
	\fontsize{7}{9}\selectfont
	\centering
	\begin{tabular}{ p{1.3cm}p{.7cm}p{.7cm}p{.7cm}p{.8cm}p{.7cm}p{.7cm} } 
		\hline
		\hline
		DMRS & DnCNN & ECNDNet & DnCNN-B & RDDCNN-B & DeamNet & DPIR \\
		\hline
		DnCNN          & 1 & 0.97 & 0.51 & 0.56 & 0.93 & 0.88\\
		ECNDNet          & 0.97 & 1 & 0.51 & 0.56 & 0.94 & 0.90 \\
		DnCNN-B    & 0.51 & 0.51 & 1 & 0.83 & 0.52 & 0.54 \\
		RDDCNN-B& 0.56 & 0.56 & 0.83 & 1 & 0.58 & 0.60 \\
		DeamNet           & 0.93 & 0.94 & 0.52 & 0.58 & 1 & 0.92 \\
		DPIR                 & 0.89 & 0.90 & 0.54 & 0.60 & 0.92 & 1 \\
		\hline
		\hline
	\end{tabular}
\label{ta:similitude}
\end{table}

\subsection{Color image denoising adversarial attack}
Under the constraint of denoising-PGD, 3-channel color RGB adversarial samples are generated for non-blind denoising model FFDNet with noise level set to 25, maximum perturbation $\varepsilon$ set to 0.055, and 5 iterative steps.

\begin{figure}[!t]
	\centering
	\includegraphics[width=1\linewidth]{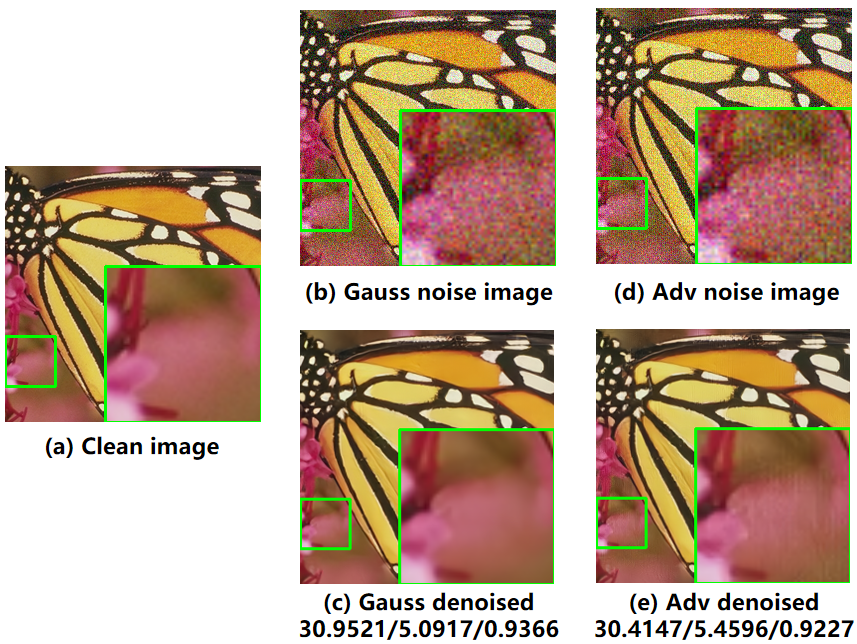}
	\caption{Effect of denoising-PGD on color RGB images (butterfly).}
	\label{f:color}
\end{figure}

As can be seen in Fig. \ref{f:color}, the addition of adversarial noise does not make a perceptible difference to the noise image, but makes the model denoised substantially less effective. We can see that PSNR decreases by 0.5374 dB, MAE increases by 0.3679, and SSIM decreases by about 0.0139. Visually, the enlarged areas in the denoised image produce the wrong texture, and many image details are erased.

In the experiments to test the transferability, we selected multiple types of deep image denoising models for color RGB image denoising: non-blind denoising methods FFDNet and BRDNet, blind denoising methods Noise2Noise and FAN, plug-and-play methods CurvPnP and DPIR. The adversarial samples generated for FFDNet under denoising-PGD are tested on the above models.

\begin{table}
\caption{Transferability of RGB adversarial samples generated by FFDNet to various models on Set3c}
\fontsize{7}{9}\selectfont
\centering
\begin{tabular}{ p{1.3cm}p{.7cm}p{.7cm}p{.7cm}p{.7cm}p{.7cm}p{.7cm} } 
	\hline
	\hline
	& Gauss & Adv & Gauss & Adv & Gauss & Adv \\
	Image & PSNR & PSNR & MAE & MAE & SSIM & SSIM \\
	\hline
	FFDNet          & 31.06 & 30.13 & 5.09 & 6.00 & 0.934 & 0.924\\
	BRDNet          & 31.22 & 30.63 & 5.29 & 5.73 & 0.938 & 0.929 \\
	Noise2Noise    & 29.24 & 28.87 & 6.46 & 6.78 & 0.913 & 0.909 \\
	FAN                  & 31.58 & 31.24 & 4.82 & 4.89 & 0.942 & 0.938 \\
	CurvPnP           & 31.68 & 30.51 & 5.02 & 5.84 & 0.944 & 0.915 \\
	DPIR                 & 31.60 & 30.49 & 5.05 & 5.90 & 0.942 & 0.911 \\
	\hline
	\hline
\end{tabular}
\label{ta:colortran}
\end{table}

As can be seen from Table \ref{ta:colortran}, the transferability of denoising-PGD on RGB images generated for FFDNet is less than the transferability on grayscale images generated for DnCNN. The adversarial examples show a high adversarial effect on the same type of denoising model, with PSNR decreasing by 0.59 dB, MAE increasing by 0.44, and SSIM decreasing by 0.01 for FFDNet. The adversarial samples show a higher adversarial effect on the plug-and-play model, with PSNR decreasing about 1.1 dB. While for blind denoising model, the adversarial effect decreases, but are still have about 0.3 dB in PSNR. Based on this, we can say that the best transferable adversarial samples can be obtained by applying the image denoising adversarial attack method to a non-blind deep image denoising model. The observation also coincides with the result in section III.C, i.e., the non-blind denoising model has the smallest adversarial space.

\subsection{Deep image denoising adversarial training}

In terms of defense, there are two main categories: adversarial defense and adversarial robustness. Adversarial defense methods usually obfuscate the loss gradient, making effective attacks harder to find, and they often succeed against specific, weaker attacks and fail against stronger ones. In this sense, networks that use adversarial defense methods are secure against weak attacks, but are not truly robust because the adversarial samples remain. Currently, the only method that has demonstrated true adversarial robustness is adversarial training, in which the network adds adversarial samples as input to the training process.

\begin{figure}[!t]
	\centering
	\includegraphics[width=1\linewidth]{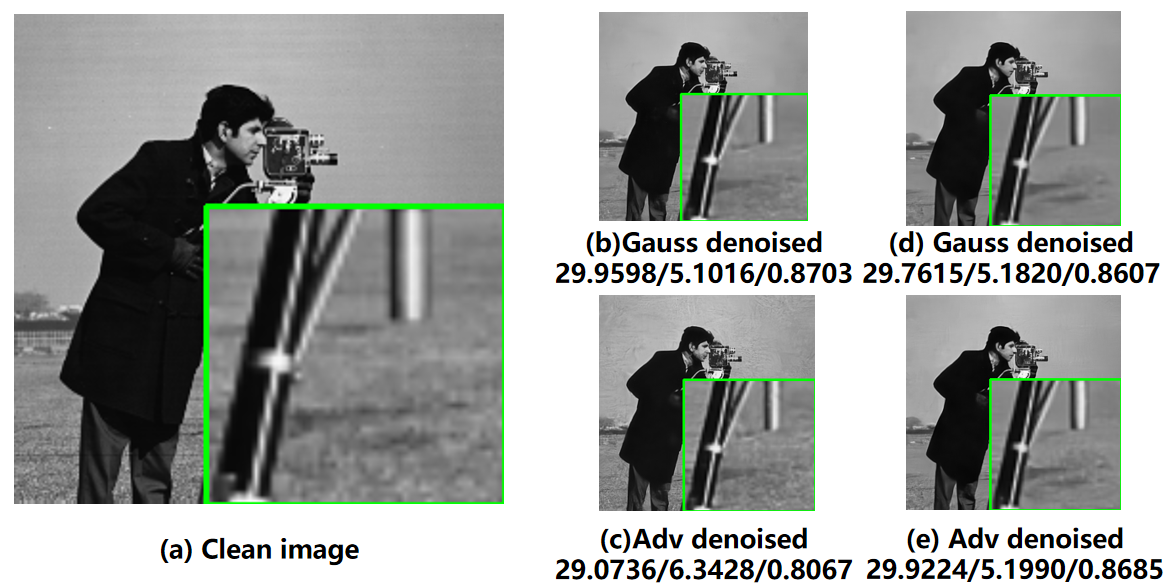}
	\caption{Comparison of the results of the adversarial training test. (a) is the clean image. (b)(c) shows the denoising effect of the original model. (d)(e) shows the denoising effect of the model after adversarial training.}
	\label{f:advtrainpsnr}
\end{figure}

\begin{figure}[!t]
	\centering
	\includegraphics[width=1\linewidth]{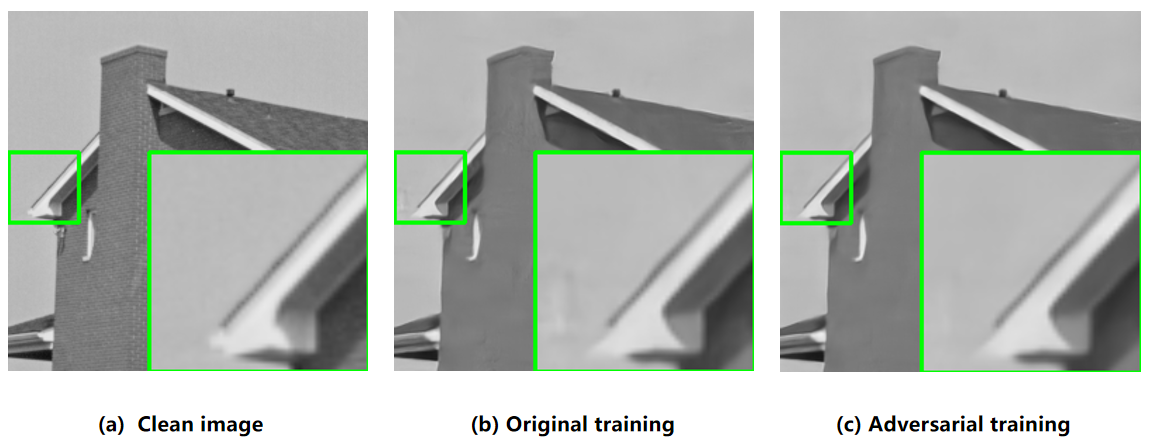}
	\caption{Denoising effect of the adversarial training model and the original model when facing the same Gaussian noise image.}
	\label{f:advtraineye}
\end{figure}

We used denoisng-PGD at an iteration number of 5 and $\varepsilon=0.012$ to generate the adversarial noise images. Mix the adversarial noise images and Gaussian noise images in 1:1 ratio and feed into the model training phase. The adversarial samples re-generated by denoising-PGD on DnCNN and DnCNN-B are integrated as a test set for testing the resistance of the model after adversarial training against adversarial samples. As can be seen in Fig. \ref{f:advtrainpsnr}, the PSNR decremental are reduced from the original 0.8862 dB to 0.1609 dB, and visually denoised adversarial samples become closer to original images. Evaluation of the effect of adversarial training is showed in Table \ref{ta:advtrain}. The difference in the average denoising effect before and after adversarial training is reduced from 0.8 dB to an impressive 0.1 dB. Surprisingly, the adversarial trained denoising model doesn’t show any performance decreasing on original Gaussian noise images, as the main drawback of adversarial training is performance decreasing on the original task. Fig. \ref{f:advtraineye} shows that the adversarial trained model has even better denoising effect on pure Gaussian noise images than vanilla deep image denoising model. The adversarial trained model can reduce artifacts which are commonly appeared in deep image denoising models.

\begin{table*}
\caption{Evaluation of the adversarial training.}
\fontsize{7}{9}\selectfont
\centering
\begin{tabular}{ llllllllllllll|l } 
	\hline
	\hline
	  &        &Cameraman&House&Pepper&Fishstar&Monarch&Airplane&Parrot&Lena&Barbara&Ship&Man&Couple&Avg.(RES)\\
	\hline
	           &PSNR&29.8&33.2&30.6&29.2&30.0&28.8&29.0&30.7&28.3&28.7&28.6&28.5&29.6 \\
	Gauss &SSIM&0.86&0.86&0.88&0.86&0.91&0.86&0.85&0.88&0.82&0.82&0.79&0.82&0.85 \\
	           &MAE&5.2&3.9&5.3&6.4&5.4&5.9&6.1&5.1&7.0&6.6&6.9&6.9&5.9\\
	           &PSNR&29.9&33.0&30.6&29.3&30.1&29.0&29.3&30.8&28.6&28.9&28.8&28.6&29.7(+0.1)\\
	Adv     &SSIM&0.87&0.86&0.87&0.87&0.91&0.87&0.85&0.88&0.83&0.82&0.81&0.83&0.86(+0.01)\\
	           &MAE&5.2&4.0&5.3&6.4&5.4&5.9&6.0&5.2&6.8&6.5&6.8&6.9&5.9(-0.0)\\
	\hline
	\hline
\end{tabular}
\label{ta:advtrain}
\end{table*}

\section{Discussion}

In this chapter, we explore in Section V-A whether generating adversarial perturbation directly to the clean image can lead to better adversarial effect. In section V-B, we demonstrate that classical model-driven denoising method is resistant to adversarial attacks. It explains why many defensive methods that try to eliminate adversarial noise in the preprocessing stage tend to use classical model-driven image denoising methods in stead of the more effective deep image denoising methods in recent years. We introduce L2-denoising-PGD in Section V-C and Section V-D to further mimic the original Gaussian noise distribution.

\subsection{Image denoising adversarial perturbation on clean images}

Adversarial perturbation is actually a noise which can be directly applying to the clean image instead of applying to Gaussian noise images as what we did in this work.  However, adversarial perturbation applied to the clean image follows different distribution instead of Gaussian and results less adversarial effect.

\begin{figure}[!t]
	\centering
	\includegraphics[width=0.98\linewidth]{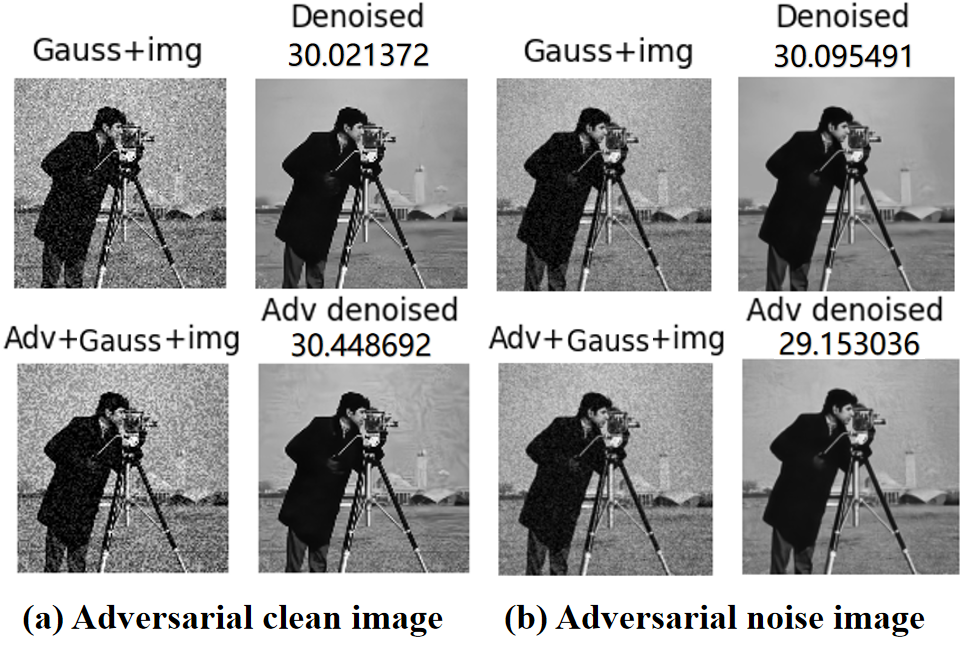}
	\caption{Demonstration of the effect of generating adversarial samples based on the clean and noise images (a) based on the clean image (b) based on the noise image.}
	\label{f:cleanvsnoise}
\end{figure}

To maintain a similar noise standard deviation, the parameter setting of 10 iterations and $\varepsilon=0.157$ was used for generating the adversarial samples on the clean image. The parameter setting of 5 iterations and $\varepsilon=0.012$ was used for generating the adversarial samples on the Gaussian noise image. As can be seen from Fig. \ref{f:cleanvsnoise}, generating adversarial attack on the clean image shows little adversarial effect with tiny PSNR drop in the denoising result, although there is more artifacts. Generating adversarial samples on noise image has obvious adversarial effect. The denoised image has texture change and new artifacts, meanwhile the denoising result PSNR decreases significantly. Thus, in this work, the adversarial perturbation is applied on the Gaussian noise images. From another aspect, the adversarial perturbation is an attack for the Gaussian noise instead of the image content, which is consistent with the phenomena that robustness is image content invariant.

In summary, the image denoising network do not have adversarial phenomenon in the case that generating adversarial instances on the clean image. We conjecture that similar to the classification boundary of image classification task, image denoising also has a denoising space, and the outide of the space is the part of the network that is difficult to denoise. Therefore, in this paper, we use the noise map with Gaussian noise added to the input to generate the adversarial samples.

\subsection{Image denoising adversarial attack on the classic method}

Among the adversarial defense methods based on denoising, most of them use classical model-driven methods as purifiers. We test the adversarial samples generated by DnCNN in the classical model-driven image denoising method BM3D.

\begin{figure}[!t]
	\centering
	\includegraphics[width=1\linewidth]{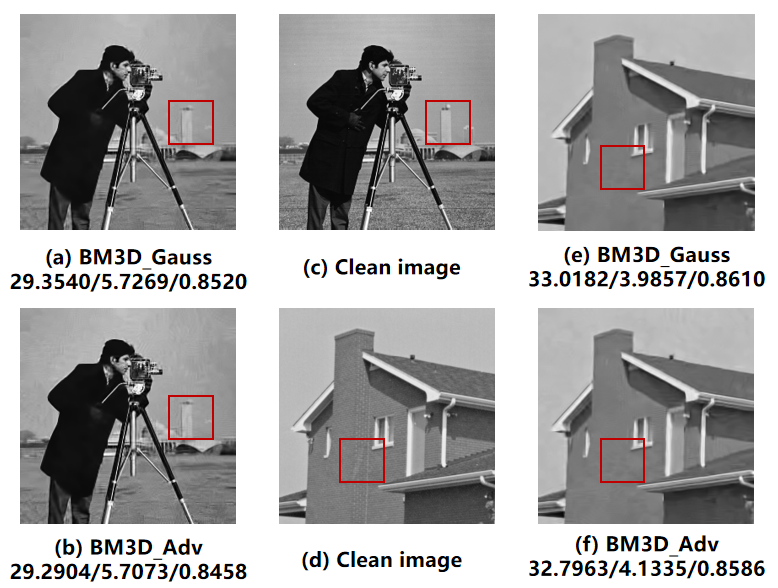}
	\caption{Effectiveness of BM3D under image denoising adversarial attacks, where the adversarial samples are generated by DnCNN under denoising-PGD. The test values are PSNR/MAE/SSIM from left to right, respectively.}
	\label{f:BM3D}
\end{figure}

As shown in Fig. \ref{f:BM3D}, the adversarial attack of image (c) has almost no effect on BM3D, and its PSNR after denoising only differs from that of the denoised Gaussian noise image by 0.0636 dB, which is within the range of normal denoising fluctuations. The denoised adversarial sample of image (d) also only differs from the denoised Gaussian noise image by 0.2219 dB in PSNR. Thus, the classical model-driven image denoising method BM3D shows strong resistance to the adversarial samples generated by DnCNN.

\subsection{L2-denoising-PGD image denoising adversarial attack}
The denoising-PGD has an uniformly distributed random perturbations on the images for generating adversarial samples before starting the iteration, it inevitably causes changes in the noise distribution. Zhu et al.[33] proposed that moving an image out of its original distribution can enhance the adversarial transferability. To ensure the experimental rigor in distribution, we impose L2 constraint to the denoising-PGD, i.e. changing the random perturbation to Gaussian, namely L2-denoisng-PGD. Algorithm 1 describes the detailed steps of the L2-denoisng-PGD.

\begin{algorithm}[H]
	\caption{L2-denoising-PGD image denoising adversarial attack.}\label{alg:alg1}
	\begin{algorithmic}
		\STATE 
		\STATE {\textsc{INPUT:}}{ Gaussian noise sample $x$; its clean image $y$; denoiser $D$; loss function $J$; perturbation size $\varepsilon$; maximum iterations $T$. } 
		\STATE {\textsc{OUTPUT:}}{ adversarial sample ${x}'$.} 
		\newline
		\STATE $ \alpha  \gets \frac{\varepsilon }{T} $
		\STATE $ {x_0}' \gets x$
		\STATE \textbf{for} $t = 0:T-1$ \textbf{do} 
		\STATE \hspace{0.4cm}{randomly Gaussian perturbations on $x$} 
		\STATE \hspace{0.4cm}{get the denoised image $D(x)$}
		\STATE \hspace{0.4cm}{calculate the gradient by $\bigtriangledown _{x} J(D(x),y)$}
		\STATE \hspace{0.4cm}$x^{t+1}=x^{t}+\alpha sign(\bigtriangledown _{x} J(D(x),y))$
		\STATE \hspace{0.4cm}$x^{t+1}=clip_{L_2} (x^{t+1}-y)+y$
		\STATE \hspace{0.4cm}$x^{t+1}=clip_{image\_range} (x^{t+1})$
		\STATE \hspace{0.4cm}$t=t+1$
		\STATE \textbf{end}
	\end{algorithmic}
	\label{alg1}
\end{algorithm}

\subsection{The effect of L2-denoising-PGD adversarial attack}
\begin{figure}[!t]
	\centering
	\includegraphics[width=1\linewidth]{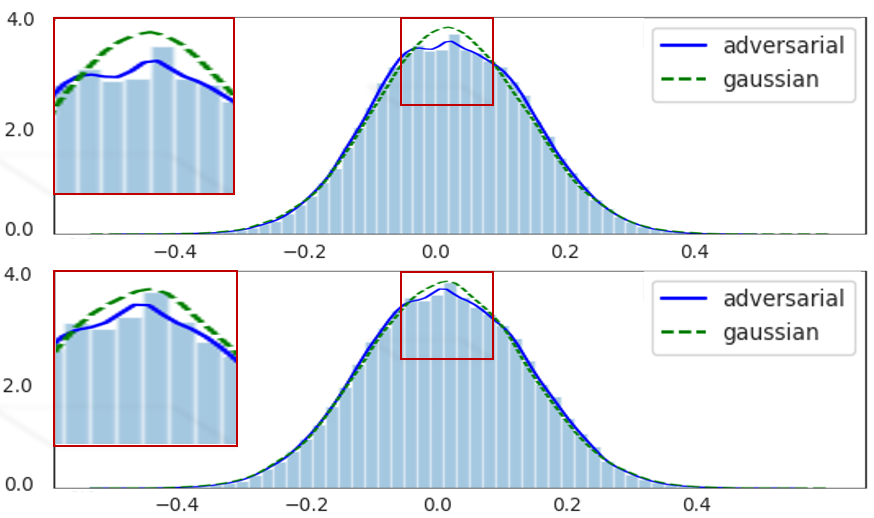}
	\caption{Visual comparison of the noise distribution of the adversarial samples generated under denoising-PGD with the Gaussian noise corresponding to the same noise level. (Upper) denoising-PGD, W-distance=0.0024. (Lower) L2-denoising-PGD, W-distance=0.0017.}
	\label{f:distribution}
\end{figure}

In order to eliminate the disturbing pitfall that the random perturbation of the PGD method may have an impact on the noise distribution, the L2-denoising-PGD image denoising adversarial attack method is proposed. As can be seen from the distribution comparison plot in Fig. \ref{f:distribution}, the adversarial noise of denoising-PGD still has some influence on the distribution, which overlaps more with the Gaussian distribution curve after the L2 constraint is applied. In other words, the noise distribution of denoising-PGD is close to the Gaussian distribution assumption, while the noise distribution of the L2-denoising-PGD is more consistent with the Gaussian distribution assumption after the constraint is applied.

\begin{figure}[!t]
	\centering
	\includegraphics[width=1\linewidth]{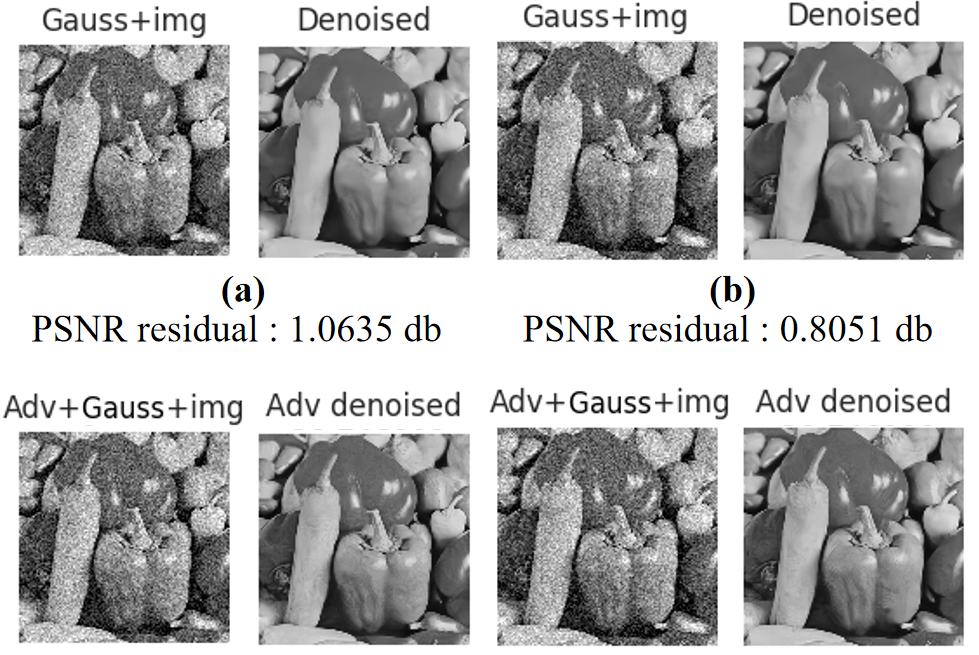}
	\caption{The effect diagram of the image denoising adversarial attack (a) denoising-PGD. (b) L2-denoising-PGD.}
	\label{f:L2show}
\end{figure}

The L2-denoising-PGD image denoising adversarial attack method also attaches an L2 constraint to the perturbation range of the noise, which further enhances the imperceptibility of the adversarial perturbation and further reduces the impact on the original image. As can be seen from Fig. \ref{f:L2show}, the effect of the adversarial under such a constraint that conforms to the Gaussian distribution noise assumption and limits the noise perturbation range is very little affected, and the PSNR residual is 0.8051 dB, which is 0.2584 dB lower than the case without the constraint, and the image still has a significant adversarial effect.

In addition, by using the L2-denoising-PGD on different kinds of deep neural denoising networks to generate grayscale adversarial samples and color adversarial samples and testing their adversarial and transferability on the remaining classes of deep neural denoising networks, we can witness that the method is still superior in maintaining the noise distribution, eliminating the concern that the adversarial effect is significantly related to the change of the noise distribution.

\section{Conclusion}
In this work, we investigate a new task called image denoising adversarial attack, aiming to explore whether deep image denoising methods are robustness to small adversarial perturbations. Based on this idea, we propose the denoising-PGD image denoising adversarial attack and the L2-denoising-PGD attack with constraints to the noise distribution. Through extensive experiments, we verify that the proposed method has consistent excellent adversarial effect as well as transferability among different types of deep image denoising models. Deep image denoising methods are not robust, especially blind denoising models. The pure data-driven non-blind denoising model have shown better robustness comparing to hybrid-driven models and non-blind denoising models. Unlike adversarial attack tasks targeting image classification, denoising-PGD shows surprising transferability in image denoising. We conjecture and verify that this is due to the high similitude of deep image denoising methods, and the adversarial region can form a continuous space. In addition, we find that the classical model-driven image denoising approach shows resistance to the adversarial attack.

\bibliographystyle{./IEEEtran}
\bibliography{./IEEEexample}

\newpage

\section{Biography Section}
\begin{IEEEbiography}[{\includegraphics[width=1in,height=1.25in,clip,keepaspectratio]{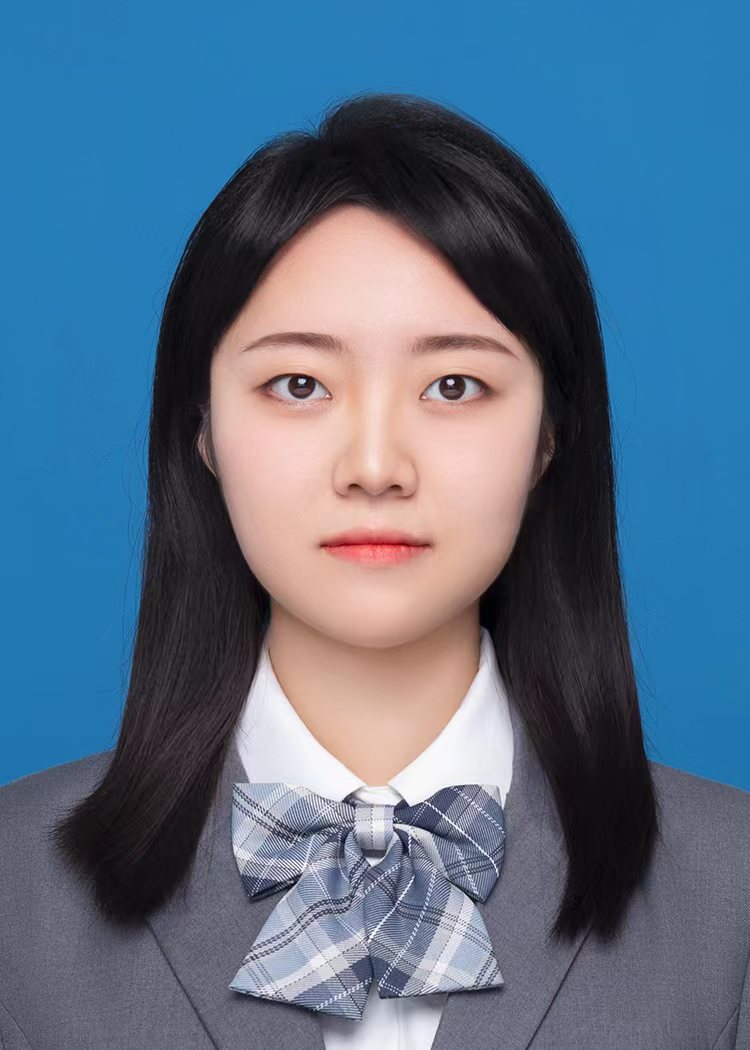}}]{Jie Ning}
Jie Ning received the B.S. degree in information and computing sciences from Ningxia University, Yinchuan, China, in 2021. She is currently working toward the Ph.D. degree with the School of Mathematics, Harbin Institute of Technology, Harbin, China. Her research interests include image restoration, partial differential equations, and deep learning.
\end{IEEEbiography}

\begin{IEEEbiography}[{\includegraphics[width=1in,height=1.25in,clip,keepaspectratio]{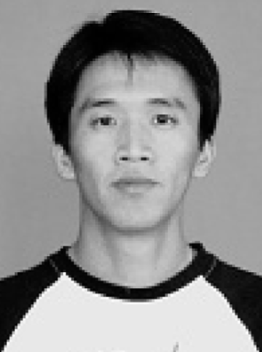}}]{Jiebao Sun}
Jiebao Sun received his Ph.D. degree from Jilin University, Changchun, China, in 2008. Subsequently, he spent two years as a Postdoctoral Researcher at the Harbin Institute of Technology, Harbin, China. He is currently a professor at School of Mathematics, Harbin Institute of Technology, Harbin, China. His research interests include partial differential equations, nonlinear diffusion, and mathematical methods in image analysis.
\end{IEEEbiography}

\begin{IEEEbiography}[{\includegraphics[width=1in,height=1.25in,clip,keepaspectratio]{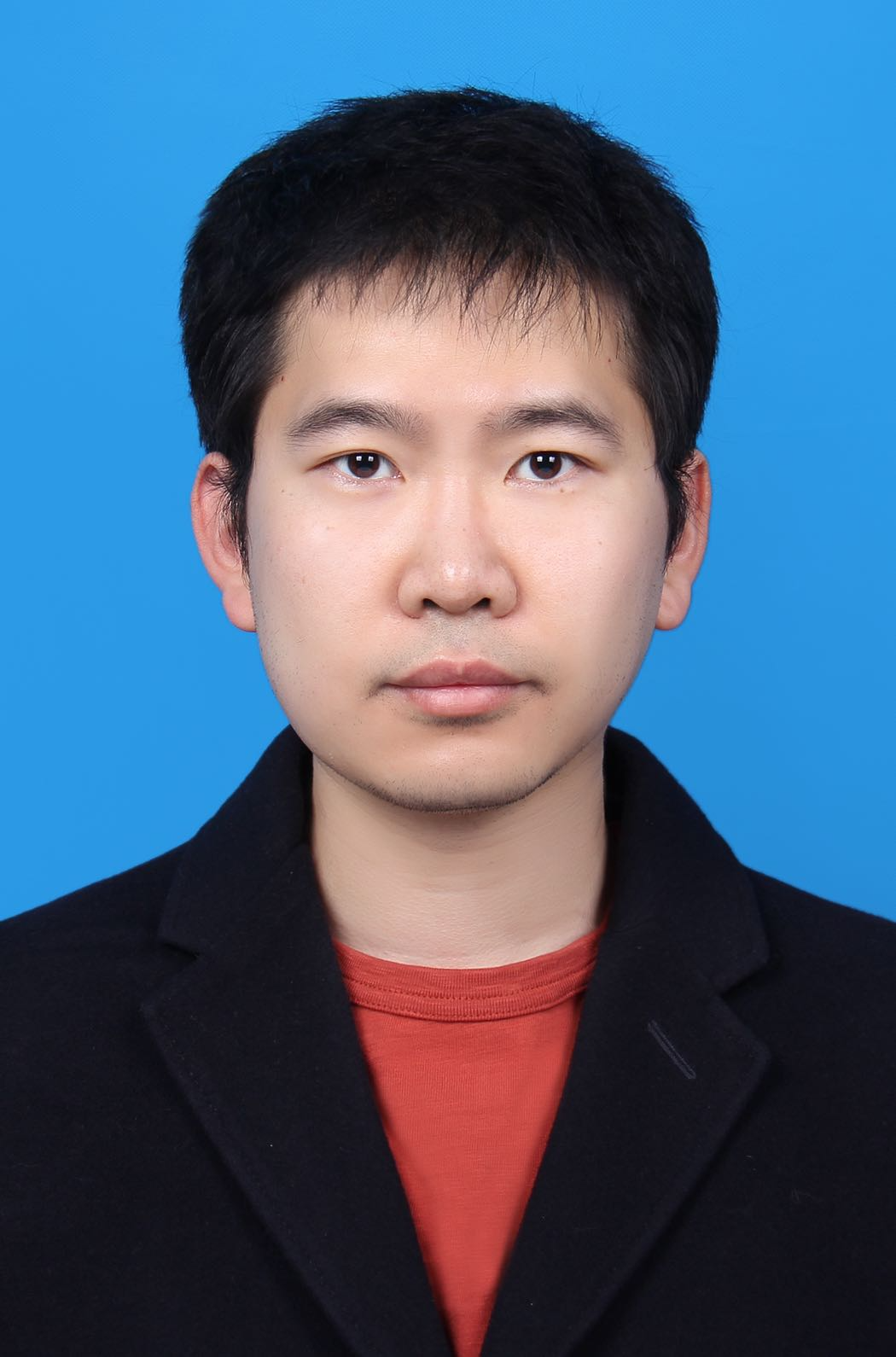}}]{Yao Li}
Yao Li received the Ph.D degree in system engineering from University of Illinois Urbana-Champaign, Illinois, US, in 2020. He is currently working as assistant professor with the School of Mathematics, Harbin Institute of Technology, Harbin. His research interests include adversarial attack, image processing, and EEG processing.
\end{IEEEbiography}

\begin{IEEEbiography}[{\includegraphics[width=1in,height=1.25in,clip,keepaspectratio]{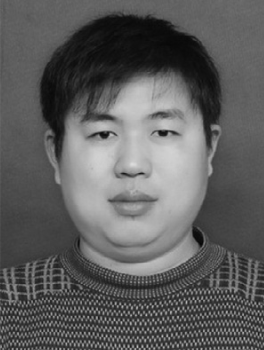}}]{Zhichang Guo}
Zhichang Guo received his Ph.D. degree from Jilin University, Changchun, China, in 2010. He is currently an associate professor at School of Mathematics, Harbin Institute of Technology, Harbin, China. His research interests include partial differential equations, nonlinear analysis, mathematical methods in image analysis, and deep learning.
\end{IEEEbiography}

\begin{IEEEbiography}[{\includegraphics[width=1in,height=1.25in,clip,keepaspectratio]{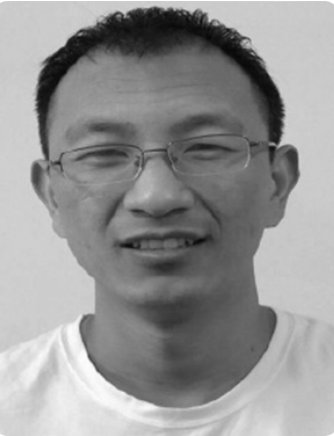}}]{Wangmeng Zuo}
Wangmeng Zuo (Senior Member, IEEE) received the Ph.D. degree from the Harbin Institute of Technology, Harbin, China, in 2007. He is currently a professor with the School of Computer Science and Technology, Harbin Institute of Technology. He has authored or coauthored more than 100 papers in top tier journals and conferences. His publications have been cited more than 30,000 times in literature. His research interests include image enhancement and restoration, image and face editing, object detection, visual tracking, and image classification. He is on the editorial boards of IEEE Transactions on Pattern Analysis and Machine Intelligence and IEEE Transactions on Image Processing.
\end{IEEEbiography}

\vfill

\end{document}